\documentclass[sigplan,nonacm,10pt]{acmart}
\usepackage{amsmath,amsfonts}
\usepackage{url}
\usepackage{verbatim}
\usepackage{graphicx}
\usepackage{subfigure}
\usepackage{caption}

\usepackage{algorithm}
\usepackage{algorithmic}
\usepackage{textcomp}
\usepackage{multirow}
\usepackage{url}
\usepackage{color}
\usepackage{pifont}
\usepackage{xcolor}
\usepackage{tabularx}
\usepackage{pifont}
\usepackage{booktabs}
\usepackage{makecell}
\usepackage{hyperref}
\usepackage[most]{tcolorbox}
\usepackage{subcaption}

\tcbset{
  myobservation/.style={
    enhanced,
    colback=gray!10!white,       
    colframe=gray!50!black,      
    arc=2mm,                     
    boxsep=0pt,                  
    left=3mm,                    
    right=3mm,                   
    top=1.5mm,                   
    bottom=1.5mm,                
    fonttitle=\normalfont\bfseries, 
    coltitle=black,              
    boxrule=0.5pt,               
    drop shadow,                 
    fonttitle=\bfseries,         
    title=Observation #1,        
    attach title to upper={\quad} 
  }
}

\newcommand{\fakepar}[1]{\vspace{.5mm}\noindent\textbf{#1.}}

\AtBeginDocument{%
  \providecommand\BibTeX{{%
    \normalfont B\kern-0.5em{\scshape i\kern-0.25em b}\kern-0.8em\TeX}}}

\setcopyright{rightsretained}

\begin{document}
\title{SparseDVFS: Sparse-Aware DVFS for Energy-Efficient Edge Inference}

\author{Ziyang Zhang}
\affiliation{%
  \institution{Politecnico di Milano}
  \country{Milan, Italy}}
\email{ziyang.zhang@polimi.it}

\author{Zheshun Wu}
\affiliation{%
  \institution{Harbin Institute of Technology}
  \country{Shenzhen, China}}
\email{wuzhsh23@gmail.com}

\author{Jie Liu}
\affiliation{%
  \institution{Harbin Institute of Technology}
  \country{Shenzhen, China}}
\email{jieliu@hit.edu.cn}

\author{Luca Mottola}
\affiliation{%
  \institution{Politecnico di Milano}
  \country{Milan, Italy}}
\email{luca.mottola@polimi.it}


\renewcommand{\shortauthors}{xxx et al.}

\begin{abstract}
Deploying deep neural networks (DNNs) on power-sensitive edge devices presents a formidable challenge. While Dynamic Voltage and Frequency Scaling (DVFS) is widely employed for energy optimization, traditional model-level scaling is often too coarse to capture intra-inference variations, whereas fine-grained operator-level scaling suffers from prohibitive performance degradation due to significant hardware switching latency. This paper presents SparseDVFS, a fine-grained, sparse-aware DVFS framework designed for energy-efficient edge inference. Our key insight is that operator sparsity is a primary metric for hardware frequency modulation. 
By distinguishing between compute-bound dense operators and memory-bound sparse operators, the system can apply specialized frequency triplets to maximize energy efficiency.
To overcome switching overheads and component interference, SparseDVFS incorporates three key innovations: (1) an \textbf{offline modeler} that established a deterministic mapping between operator sparsity and optimal frequency triplets (CPU/GPU/EMC) via white-box timeline analysis; (2) a \textbf{runtime graph partitioner} that utilizes a greedy merging heuristic to aggregate operators into super-blocks, balancing scaling granularity and DVFS switching latency through a latency amortization constraint; and (3) a \textbf{unified co-governor} that employs a frequency unified scaling engine (FUSE) and a look-ahead instruction queue to eliminate antagonistic effects between independent controllers and hide hardware transition latencies. Extensive evaluations show that SparseDVFS achieves an average 78.17\% energy efficiency gain over state-of-the-art solutions while maintaining a superior 14\% cost-gain ratio.
\end{abstract}

\maketitle

\section{Introduction}\label{sec:introduction} 
The rapid proliferation of deep neural networks (DNNs) has fundamentally transformed modern computing, driving an unprecedented demand for high-performance inference at the network edge~\cite{zhang2023pos,nigade2022jellyfish,han2022microsecond,kim2023dream,fan2023sparse,liu2022veltair}. As these models grow in complexity, spanning from deep convolutional neural networks (CNNs) like ResNet~\cite{he2016deep} to intricate Transformer-based architectures like vision transformers (ViT)~\cite{vaswani2017attention,dosovitskiy2020image}, deploying them on resource-constrained edge devices presents a formidable challenge. Edge devices (e.g., NVIDIA Jetson series~\cite{nvidia_jetson_modules}) widely used in autonomous systems~\cite{shi2022vips,padmanabhan2023gemel,ouyang2022cosmo} and IoT applications~\cite{wen2023adaptivenet,lin2020mcunet}, operate under strict power envelopes and thermal constraints. Consequently, energy efficiency has emerged as the paramount design criterion, necessitating sophisticated system-level power management strategies.

Dynamic voltage and frequency scaling (DVFS)~\cite{pillai2001real,kim2022ztt, zhang2025e4,wang2025using,tian2025clone,lin2023workload} stands as the ubiquitous standard for managing power and performance trade-offs in modern processors. By dynamically adjusting the operating frequency and voltage of the CPU, GPU, and memory subsystems, systems can theoretically align power consumption with the instantaneous computational demands of the workload. In the context of edge AI, effective DVFS is critical. However, since traditional DVFS cannot accurately perceive DNN characteristics (such as sparsity~\cite{tang2024quest,lin2023efficient,song2024powerinfer,fan2023sparse}), leading to suboptimal frequency allocation that either causes performance degradation or significant energy waste.

\begin{figure}[tb]
\large
\centerline{\includegraphics[width=\linewidth]{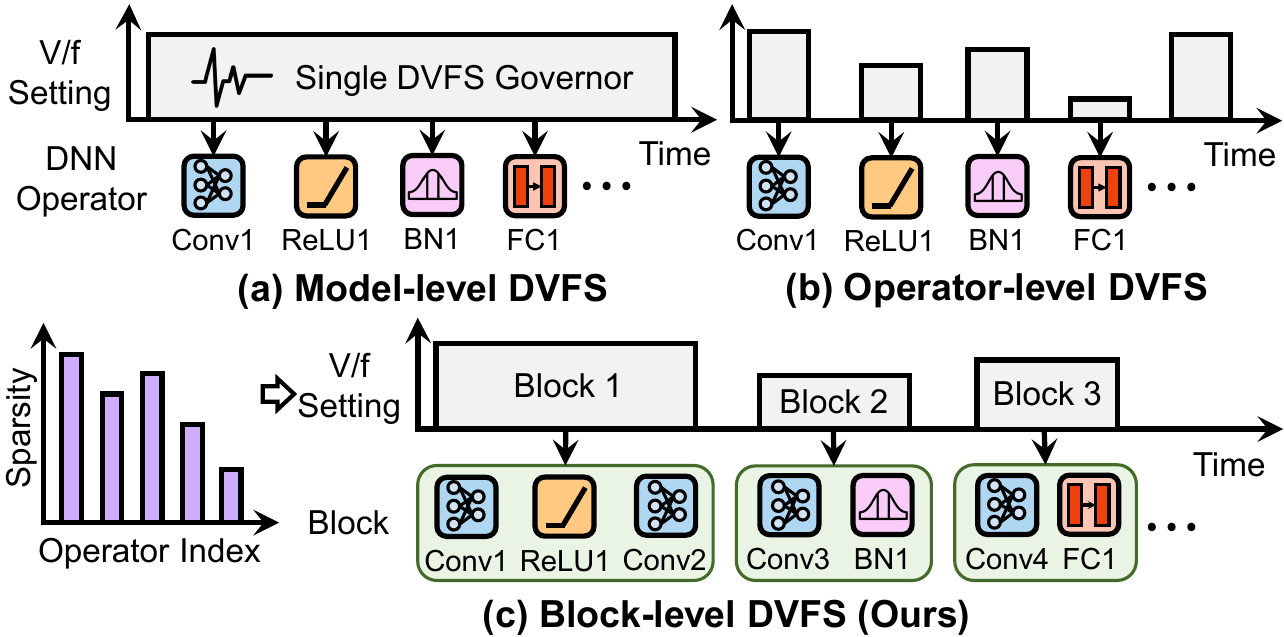}}
\caption{Comparison of DVFS governor granularities for edge DNN inference. (a) Model-level DVFS governor employs a single V/f setting. (b) Operator-level fine-grained DVFS governor. (c) Our sparsity-aware block-level DVFS governor.}
\Description{Comparison of DVFS governor granularities for edge DNN inference. (a) Model-level DVFS governor employs a single V/f setting. (b) Operator-level fine-grained DVFS governor. (c) Our sparsity-aware block-level DVFS governor.}
\label{fig:1}
\end{figure}


Existing research has proposed specific DVFS governors~\cite{kim2021ztt, lin2023workload,wang2025using,zhang2025e4,zhang2024dvfo,guerreiro2018gpgpu,kandiah2021accelwattch} for on-device DNN inference to improve energy efficiency. However, as illustrated in Figure~\ref{fig:1}, these solutions face a fundamental trade-off between scaling granularity and system-level overhead. 
Model-level DVFS in Figure~\ref{fig:1}(a) commonly employed by standard Linux governors~\cite{mercati2014linux} and learning-based methods like zTT~\cite{kim2021ztt} or GearDVFS~\cite{lin2023workload}, assigns a static frequency for the entire inference. This is inherently too coarse to capture intra-model workload fluctuations where computational intensity varies significantly between layers. Conversely, while operator-level DVFS like Ascend-DVFS~\cite{wang2025using} in Figure~\ref{fig:1}(b) attempts to track these variations at the finest grain, it remains practically infeasible on commercial hardware. Our benchmarks in Section~\ref{ob4} reveal that frequency switching latencies can exceed the execution time of lightweight operators, causing transition penalties to dominate the timeline and negate energy savings. To bridge this gap, Our proposed SparseDVFS in Figure~\ref{fig:1}(c) introduces a block-level DVFS strategy. By aggregating operators into sparsity-aware super-blocks, our framework amortizes hardware switching costs while adapting to fine-grained model features, achieving a Pareto-superior trade-off between energy efficiency and latency.

Achieving such a fine-grained, sparse-aware DVFS framework involves the following three key challenges:

\begin{itemize}
    \item \textbf{DVFS adjustment overhead:} transitioning from coarse-grained model-level scaling to fine-grained operator-level control implies frequent frequency switching. Our benchmarks in Section~\ref{ob4} reveal that GPU frequency switching latency for on-device DNN inference mostly ranges from 5ms to 10ms, but exceeds 20ms at extremely low frequencies. Since many lightweight DNN operators (e.g., ReLU~\cite{glorot2011deep,krizhevsky2012imagenet} and GELU~\cite{hendrycks2016gaussian}) only take a few milliseconds to execute, operator-level DVFS strategy would cause the switching overhead to exceed the execution time, drastically increasing end-to-end latency.
    \item \textbf{Balancing latency and power:} the trade-off between latency and power is non-linear and governed by the Roofline Model~\cite{williams2009roofline,hill2019gables,ding2019instruction,siracusa2020cad}. Our benchmarks in Section~\ref{roofline} reveal that for compute-bound operators (e.g., Conv2d~\cite{lecun1998gradient}), higher frequencies yield linear speedups. For memory-bound sparse operators (e.g., LayerNorm~\cite{ba2016layer}), however, the performance bottleneck shifts to memory bandwidth. In these cases, increasing GPU frequency beyond the ridge point primarily increases power consumption with diminishing energy efficiency gains.
    \item \textbf{Architectural adaptability and Scalability:} modern edge AI workloads exhibit extreme diversity~\cite{cui2022dvabatch,choi2022serving}, spanning diverse architectures from traditional CNNs with high intra-model sparsity to emerging Transformer-based models characterized by dynamic attention patterns. Existing learning-driven DVFS governors, however, treat DNN models as black boxes, necessitating prohibitive offline retraining and manual profiling whenever a new model architecture or hardware platform is introduced. This lack of architectural transparency and deterministic modeling renders current solutions difficult to scale across the rapidly evolving landscape of DNN topologies.
\end{itemize}

To address the aforementioned limitations, this paper presents SparseDVFS, a fine-grained, sparse-aware DVFS for energy-efficient edge inference, predicated on the insight that operator sparsity is a primary metric for hardware frequency modulation.
SparseDVFS overcomes the deficiencies of existing DVFS governors through three tightly coupled components. First, the \textbf{offline modeler} employs white-box timeline analysis~\cite{henard2016comparing} and thermal-aware power modeling~\cite{kim2021ztt} to construct a deterministic mapping between operator characteristics and optimal hardware states. Second, the \textbf{runtime graph partitioner} utilizes a greedy algorithm to dynamically aggregate operators into super-blocks~\cite{ling2022blastnet}, enforcing a latency amortization constraint that ensures computational gains are not offset by transition penalties. Third, the \textbf{unified co-governor} implements a frequency unified scaling engine (FUSE) strategy~\cite{zhang2024dvfo,zhang2025e4} to synchronize CPU, GPU, and memory frequencies, effectively eliminating the antagonistic effects~\cite{zhang2025dissecting} inherent in independent component controllers. Experimental evaluations on the NVIDIA Jetson Orin Nano show that SparseDVFS offers an average energy efficiency improvement of 78.17\% and a cost gain of 14\% compared to state-of-the-art methods.

This paper makes the following contributions:

\begin{enumerate}
\item
We propose SparseDVFS, a sparse-aware block-level DVFS that bridges the gap between coarse model-level and prohibitive operator-level scaling.
\item 
We develop an white-box offline modeler that provides a deterministic mapping between operator sparsity and optimal hardware states.
\item 
We introduce a runtime graph partitioner that greedily aggregates operators into super-blocks to amortize frequency adjustment overhead.
\item 
We design a unified co-governor to eliminate the antagonistic effects of independent governors.
\end{enumerate}

\section{Motivation} \label{motivation}
SparseDVFS is motivated by five key observations regarding the interaction between DNN operators and edge hardware.
We use an NVIDIA Jetson Orin Nano with 8GB DRAM as our benchmark platform. Additionally, ResNet-18/101~\cite{he2016deep} and ViT-B16/L16~\cite{dosovitskiy2020image} are chosen as workloads.

\subsection{Roofline Model Analysis} \label{roofline}
\begin{figure}
\setlength{\abovecaptionskip}{0pt}
\setlength{\belowcaptionskip}{0pt}
\centering
\subfigure[ResNet-18]{
\begin{minipage}[b]{0.485\linewidth}
\includegraphics[width=1\linewidth]{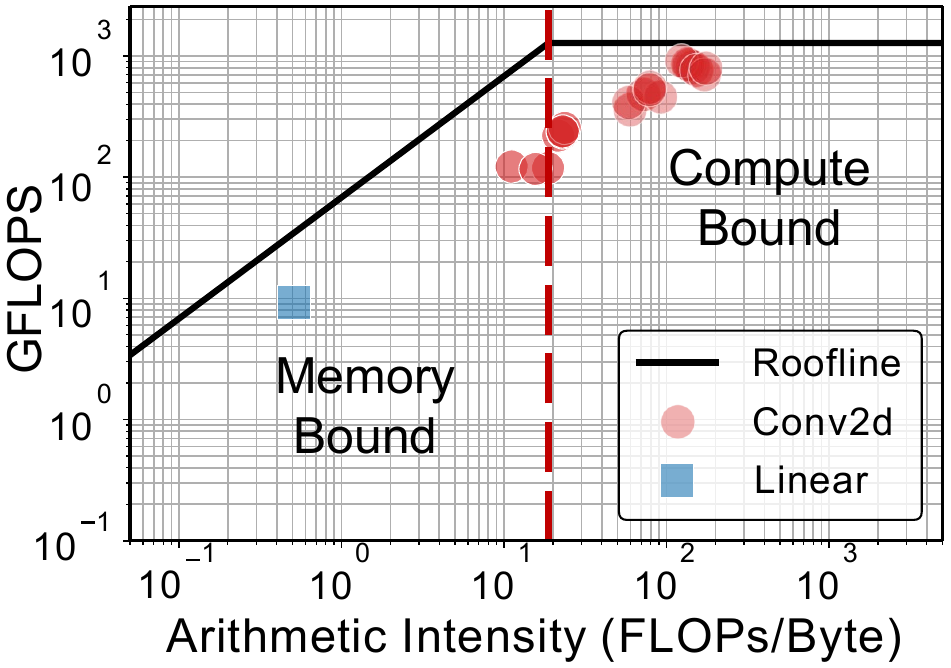}
\end{minipage}}
\subfigure[ResNet-101]{
\begin{minipage}[b]{0.485\linewidth}
\includegraphics[width=1\linewidth]{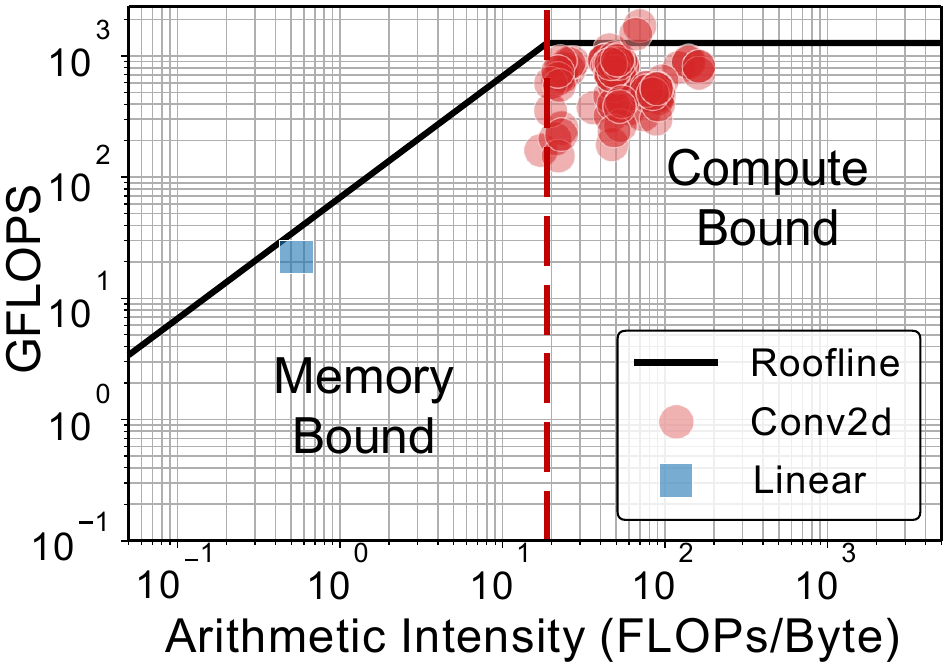}
\end{minipage}}
\subfigure[ViT-B16]{
\begin{minipage}[b]{0.485\linewidth}
\includegraphics[width=1\linewidth]{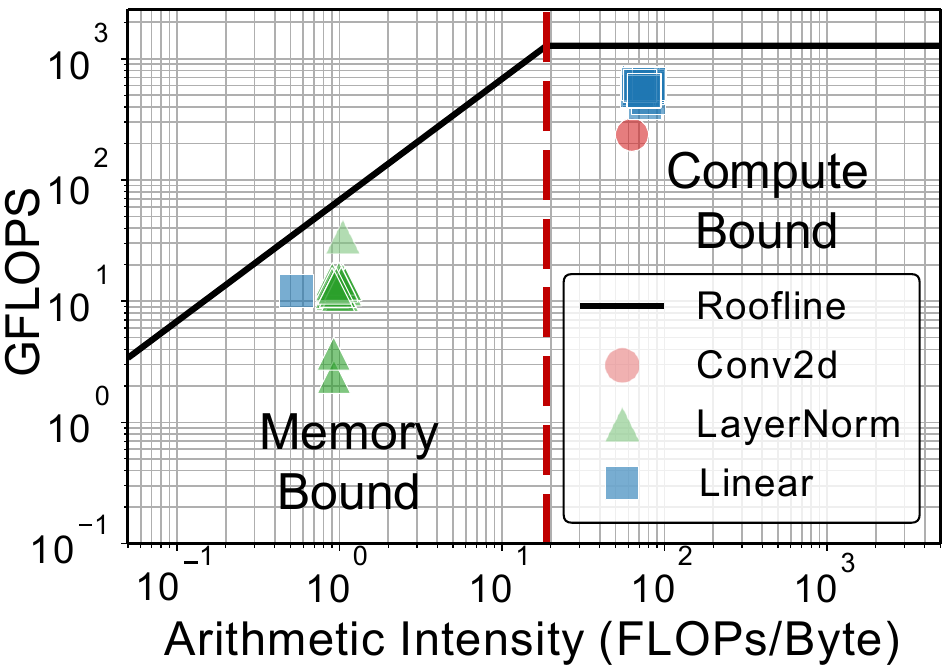}
\end{minipage}}
\subfigure[ViT-L16]{
\begin{minipage}[b]{0.485\linewidth}
\includegraphics[width=1\linewidth]{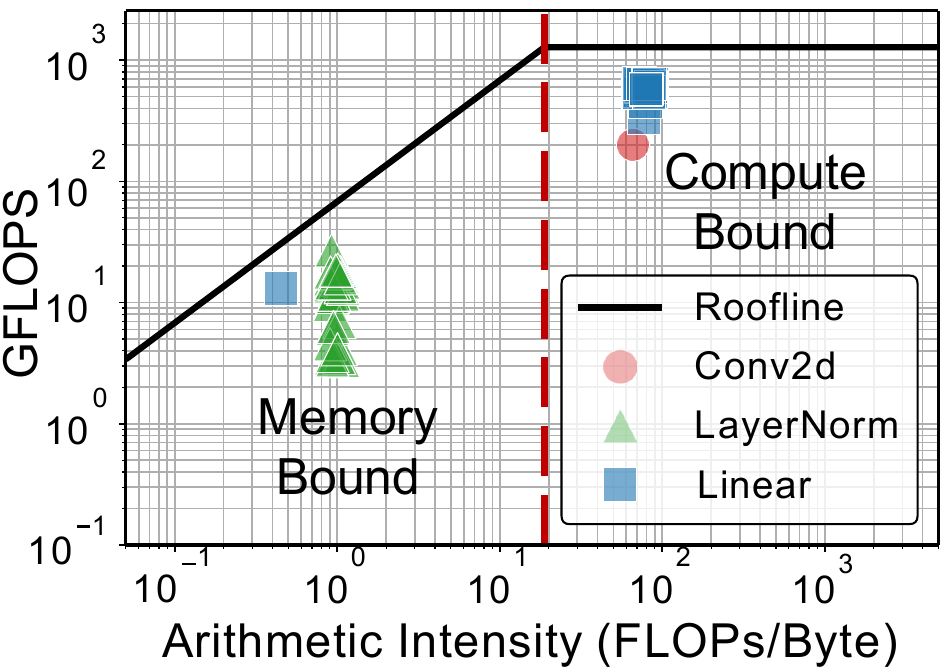}
\end{minipage}}
\caption{Roofline model across diverse DNN architectures. For CNN-based models (i.e., ResNet-18/101), dense operators (e.g., Conv2d, Linear) reside on the horizontal plateau (i.e., Compute-Bound), while activation and normalization layers fall into the sloped region (i.e., Memory-Bound).
For Transformer-based models (i.e., ViT-B16/L16), dense operators (e.g., Conv2d) reside on the horizontal plateau (i.e., Compute-Bound), while linear and normalization layers fall into the sloped region (i.e., Memory-Bound).}
\Description{Roofline model across diverse DNN architectures. For CNN-based models (i.e., ResNet-18/101), dense operators (e.g., Conv2d, Linear) reside on the horizontal plateau (i.e., Compute-Bound), while activation and normalization layers fall into the sloped region (i.e., Memory-Bound).
For Transformer-based models (i.e., ViT-B16/L16), dense operators (e.g., Conv2d) reside on the horizontal plateau (i.e., Compute-Bound), while linear and normalization layers fall into the sloped region (i.e., Memory-Bound).}
\label{fig:motivation1}
\end{figure}

The Roofline Model~\cite{williams2009roofline} provides a fundamental characterization of how different operators interact with the hardware's compute and memory limits. Figure~\ref{fig:motivation1} reveals the performance (FLOPS)~\cite{howard2017mobilenets,simonyan2014very,lecun1989backpropagation} against arithmetic intensity (FLOPs/Byte)~\cite{rogers2024aio,reggiani2022bison,oh2021layerweaver} for various operators in ResNet-18 and ViT-B16.1 The results reveal a clear bifurcation. Dense operators like Conv2d and Linear layers reside on the horizontal plateau of the Roofline, indicating they are Compute-Bound. Their performance scales linearly with frequency.

In contrast, activation functions (ReLU, GELU) and normalization layers typically fall into the sloped region, identifying them as Memory-Bound. Crucially, as the GPU frequency increases from 306 MHz to 624.75 MHz, the ridge point (where the memory-bound slope meets the compute-bound plateau) shifts. For these memory-bound operators, increasing frequency beyond a certain point yields diminishing returns in performance because the bottleneck is memory bandwidth, not compute cycles. Thus, running high-sparsity, memory-bound operators at peak frequency wastes significant energy.

\subsection{Dynamic Sparsity Distribution}
\begin{figure}
\setlength{\abovecaptionskip}{0pt}
\setlength{\belowcaptionskip}{0pt}
\centering
\subfigure[ResNet-18 (ReLU)]{
\begin{minipage}[b]{0.485\linewidth}
\includegraphics[width=1\linewidth]{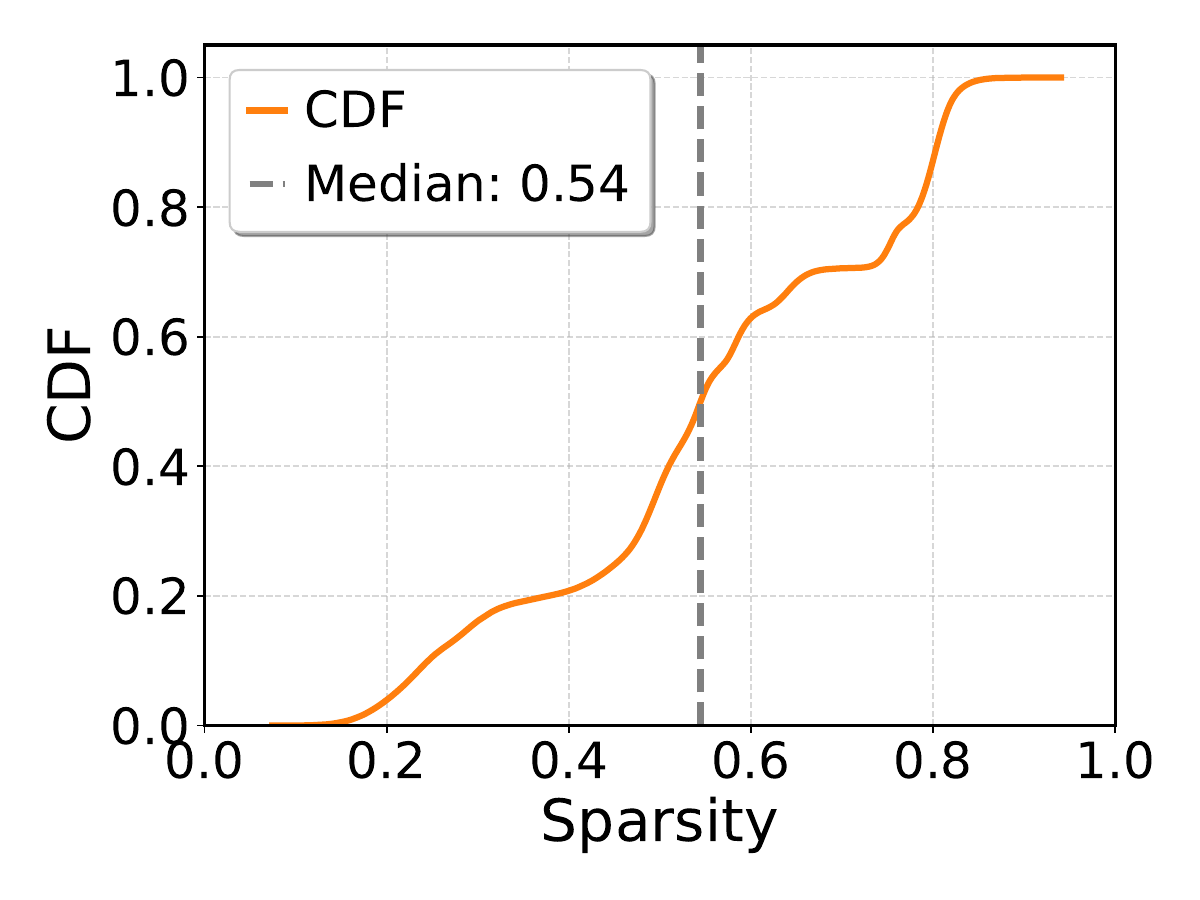}
\end{minipage}}
\subfigure[ResNet-101 (ReLU)]{
\begin{minipage}[b]{0.485\linewidth}
\includegraphics[width=1\linewidth]{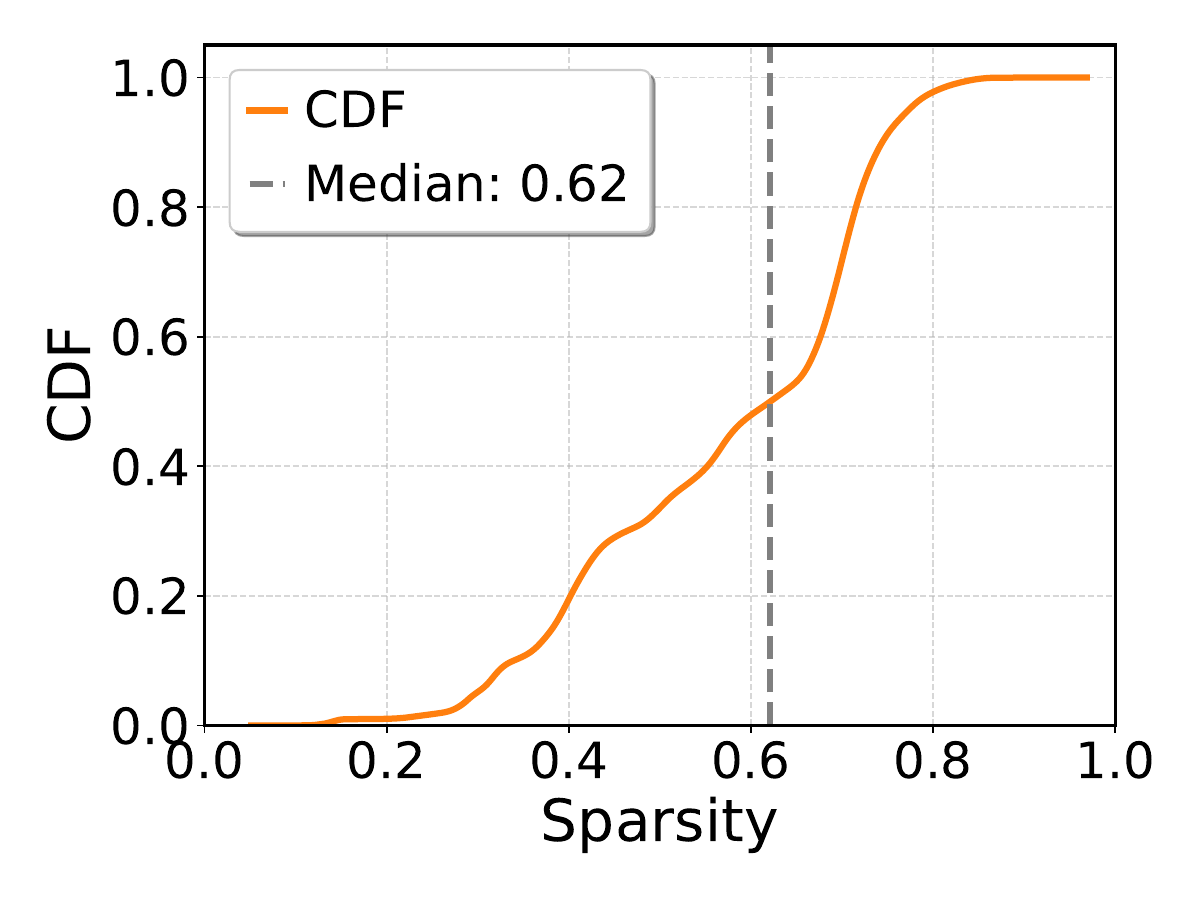}
\end{minipage}}
\subfigure[ViT-B16 (GELU)]{
\begin{minipage}[b]{0.485\linewidth}
\includegraphics[width=1\linewidth]{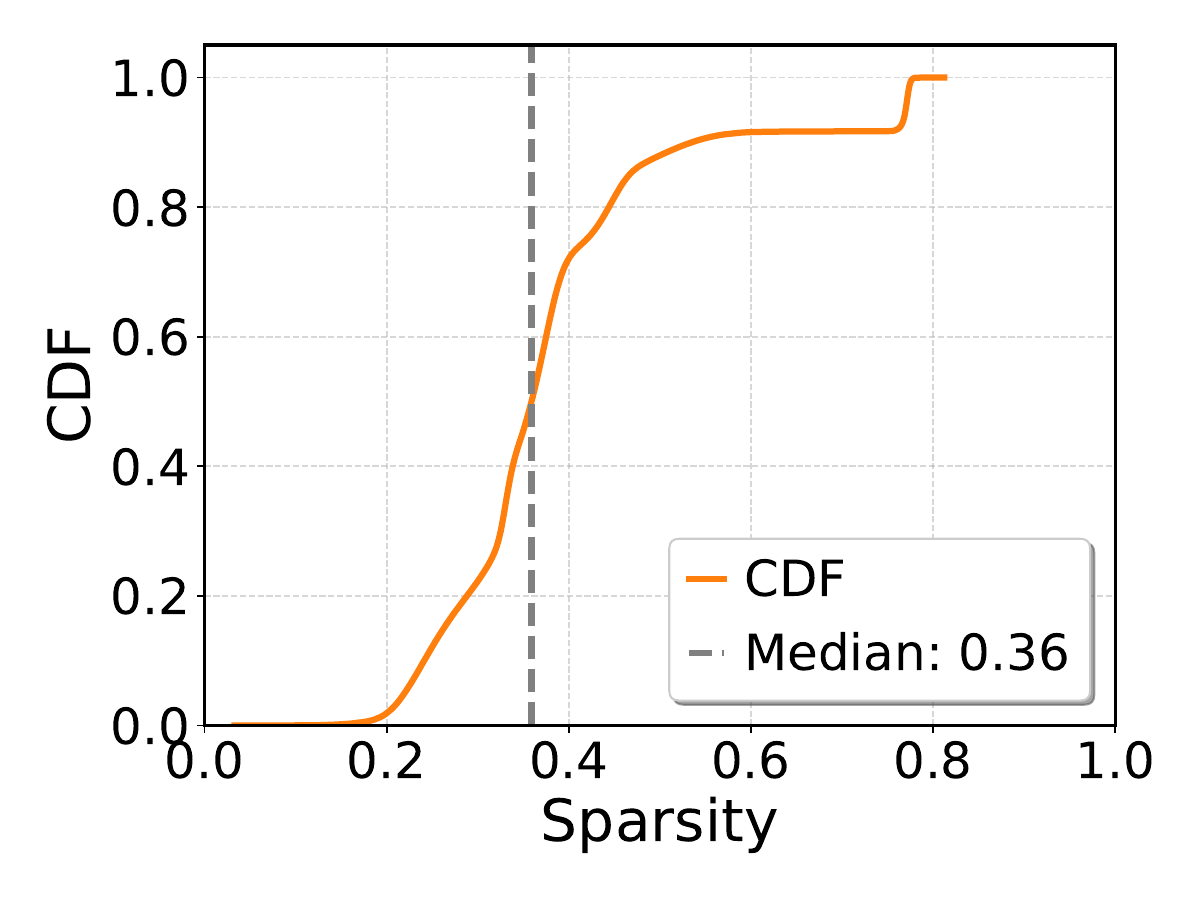}
\end{minipage}}
\subfigure[ViT-L16 (GELU)]{
\begin{minipage}[b]{0.485\linewidth}
\includegraphics[width=1\linewidth]{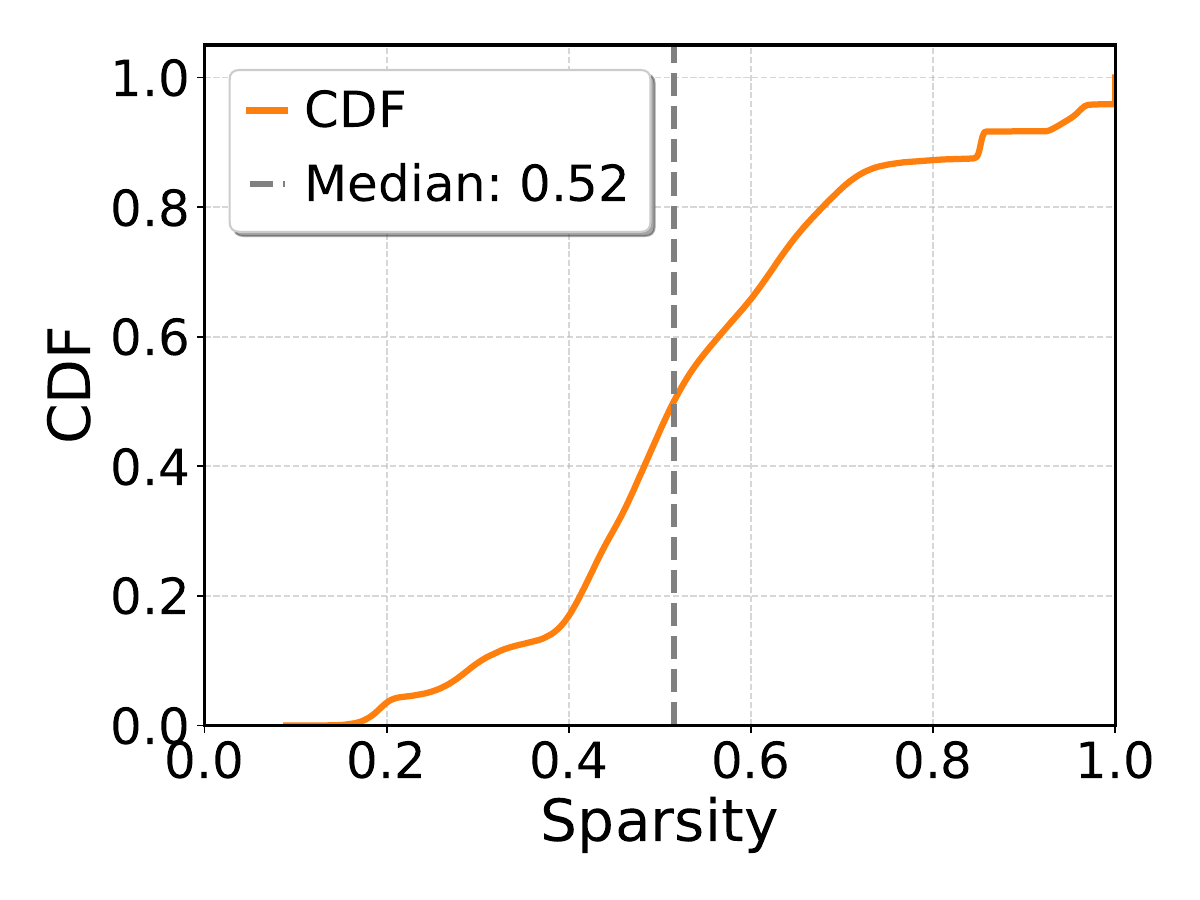}
\end{minipage}}
\caption{Dynamic sparsity distribution (CDF) for CNN and Transformer models on  ImageNet-2012 validation dataset (about 50k images). A substantial portion of operators in ResNet and ViT exhibit high sparsity (>50\%).}
\Description{Dynamic sparsity distribution (CDF) for CNN and Transformer models on ImageNet-2012 validation dataset (about 50k images). A substantial portion of operators in ResNet and ViT exhibit high sparsity (>50\%).}
\label{fig:motivation2}
\end{figure}

We analyzed the sparsity distribution of operators across different models on  ImageNet-2012 validation dataset (about 50k images)~\cite{krizhevsky2012imagenet} using cumulative distribution function (CDF).
As shown in Figure~\ref{fig:motivation2}, the CDFs for ResNet-18/101 (ReLU) and ViT-B16/L16 (GELU) show distinct profiles. For ResNet, a substantial portion of operators exhibit high sparsity (>50\%), with the median sparsity often exceeding 0.5. ViT models show a different but equally dynamic distribution, with sparsity varying significantly between Attention and MLP blocks.

In summary, the sparsity of activation functions changes dynamically with input data~\cite{fan2023sparse}. A static frequency setting cannot accommodate this variance. The wide distribution confirms that sparsity is a viable, high-entropy signal for modulating hardware performance. High-sparsity operators (right tail of the CDF) are prime candidates for frequency reduction, while low-sparsity operators (left tail) require high performance states.

\subsection{CPU-GPU Antagonistic Effect}  \label{Antagonistic}
\begin{figure}[tb]
\large
\centerline{\includegraphics[width=\linewidth]{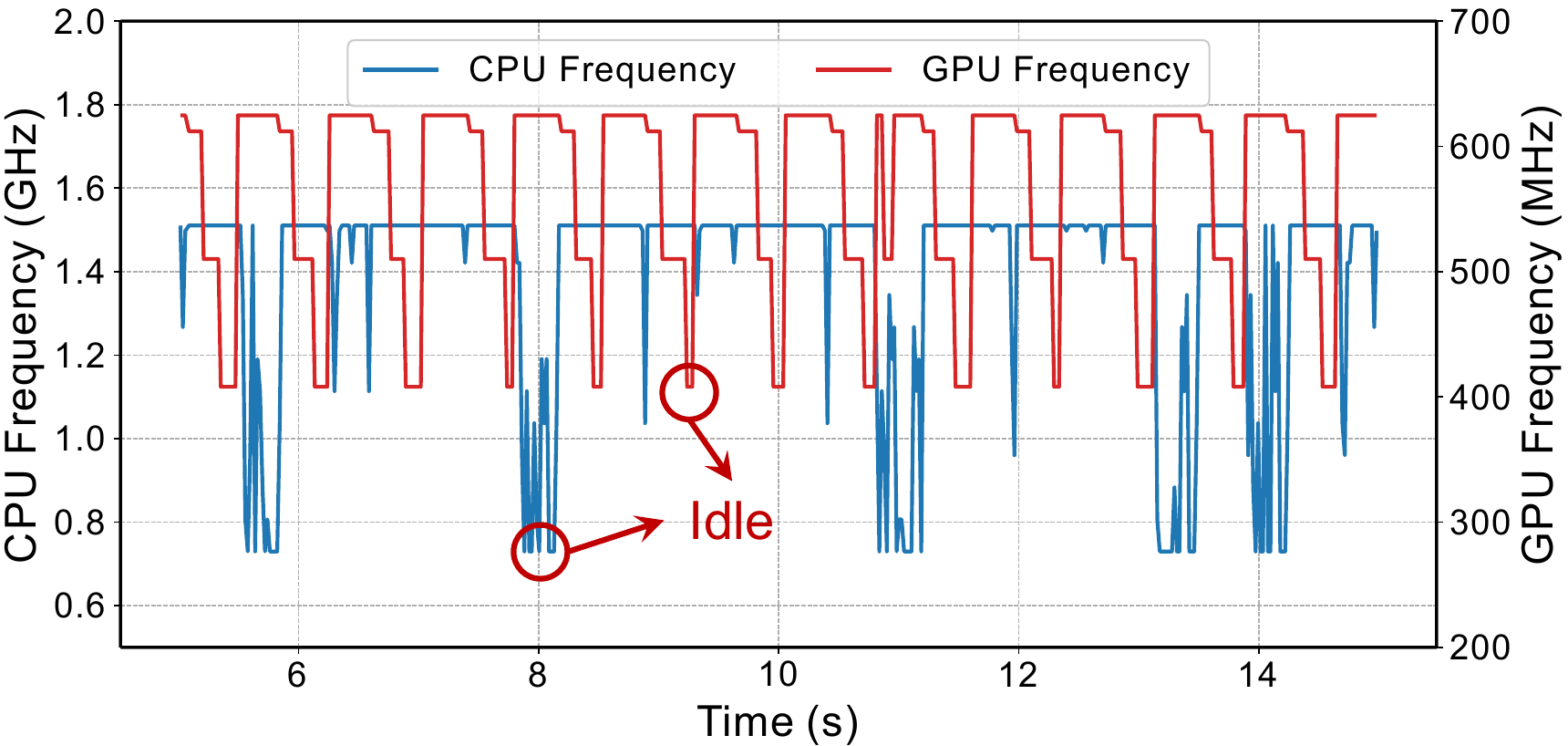}}
\caption{Runtime traces of CPU and GPU frequencies under default DVFS governors for ResNet-18. Independent and often inverse fluctuations between frequencies create an antagonistic effect.}
\Description{Runtime traces of CPU and GPU frequencies under default DVFS governors for ResNet-18. Independent and often inverse fluctuations between frequencies create an antagonistic effect.}
\label{fig:motivation3}
\end{figure}

Runtime traces of CPU and GPU frequencies under default governors (i.e., \textit{simple\_ondemand}~\cite{pallipadi2006ondemand} for GPU and \textit{Linux schedutil}~\cite{gouicem2020fewer} for CPU) reveal a critical inefficiency.
In Figure~\ref{fig:motivation3}, CPU and GPU frequencies fluctuate independently and often inversely. For instance, during a GPU-intensive computation phase, the CPU utilization drops, causing the default  governor to lower the CPU frequency. However, when the GPU finishes and requests more data, the CPU is in a low-power state, causing a latency spike while it ramps up.

This antagonistic effect~\cite{zhang2025dissecting} leads to pipeline stalls and energy wastage. To this end, a unified governor is required to synchronize CPU and GPU transitions, ensuring that the CPU is ready to feed the GPU (i.e., prefill) exactly when needed, regardless of instantaneous utilization.

\subsection{DVFS Switching Overhead vs. Inference Latency} 
\label{ob4}
\begin{figure}[tb]
\large
\centerline{\includegraphics[width=\linewidth]{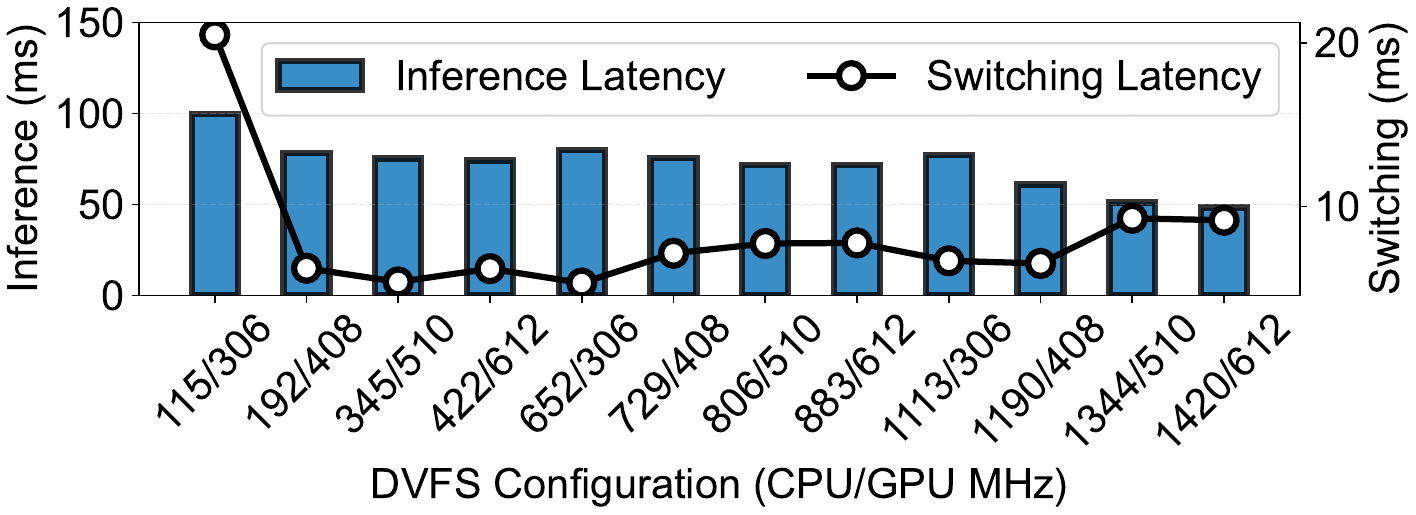}}
\caption{Comparison between DVFS switching overhead and end-to-end inference latency for ResNet-18.}
\Description{Comparison between DVFS switching overhead and end-to-end inference latency for ResNet-18.}
\label{fig:motivation4}
\end{figure}

We benchmark the latency overhead of frequency switching using DVFS under different CPU-GPU frequency configurations, and compare it against the operator execution time. As shown in Figure~\ref{fig:motivation4}, the results reveal that the DVFS switching latency mostly ranges from 5ms to 10ms. In contrast, a full inference takes between 60ms and 100ms depending on the CPU-GPU clock frequency configurations. However, the cumulative overhead of switching hardware states for every single operator would be prohibitive.
Therefore, the non-negligible nature of relative to full inference latency renders an operator-level DVFS strategy practically infeasible. Attempting to adjust frequencies for every layer would cause the switching penalty to dominate the execution timeline, potentially doubling the end-to-end latency and negating any energy savings achieved through frequency scaling.

Compared to the huge switching overhead of operator-level DVFS and the lack of awareness of DNN characteristics in model-level DVFS, block-level DVFS in SparseDVFS aggregates operators into super-blocks through the sparsity of DNNs, enabling block execution time significantly to exceed frequency switching overhead, thereby effectively amortizing hardware switching costs while capturing fine-grained features of the model.

\subsection{Impact of Clock Frequency on Performance}
\begin{figure}[tb]
\setlength{\abovecaptionskip}{0pt}
\setlength{\belowcaptionskip}{0pt}
\centering
\subfigure[Inference Latency]{
\begin{minipage}[b]{0.4825\linewidth}
\includegraphics[width=1\linewidth]{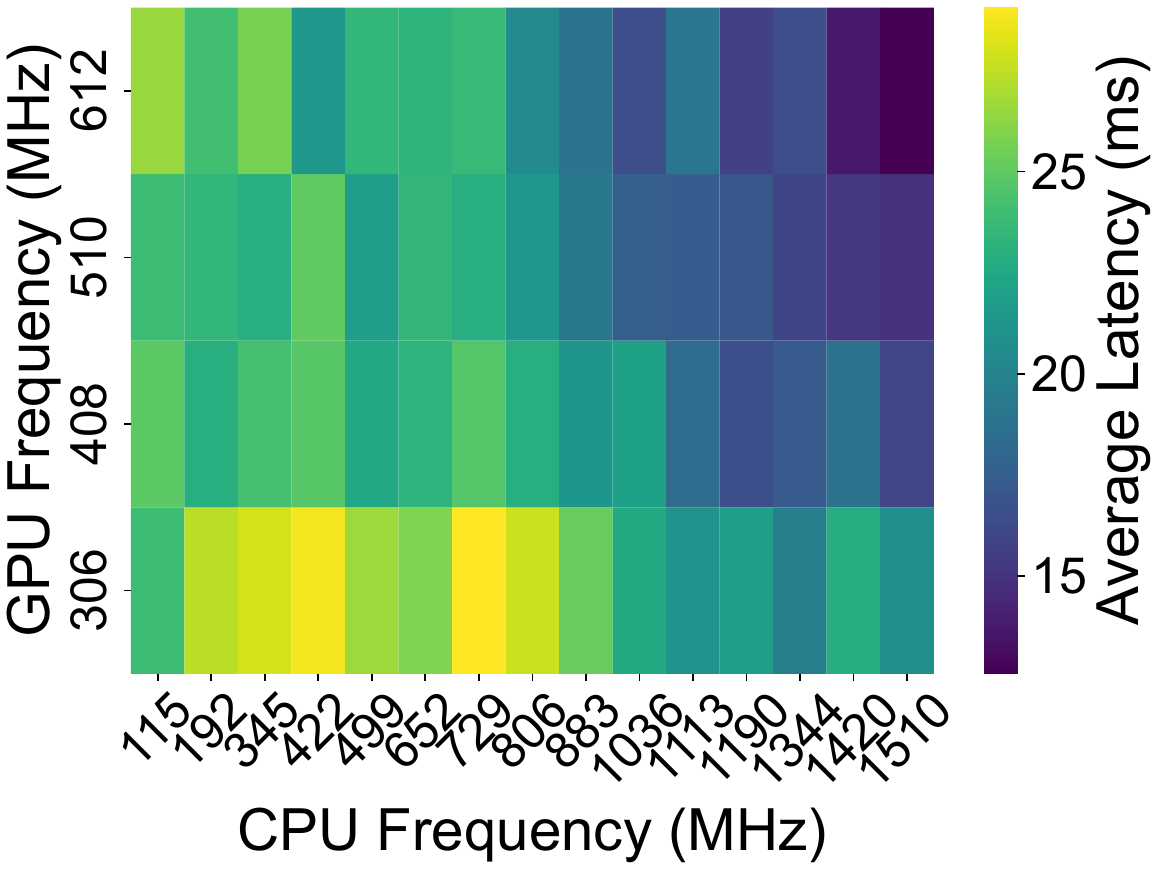}
\end{minipage}}
\subfigure[Power Consumption]{
\begin{minipage}[b]{0.4825\linewidth}
\includegraphics[width=1\linewidth]{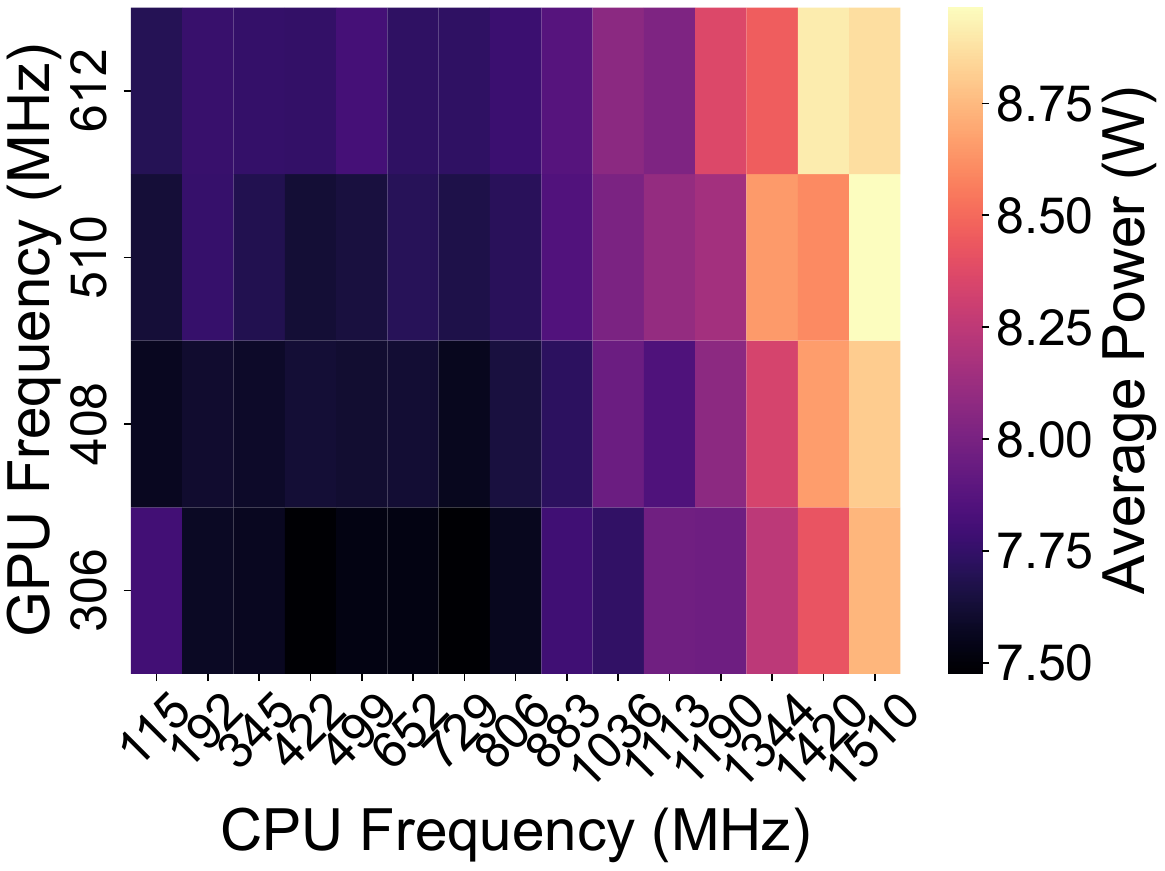}
\end{minipage}}
\caption{The impact of CPU and GPU clock frequencies on system performance for ResNet-18. The heatmaps illustrate the non-linear sensitivity of (a) inference latency and (b) power consumption to frequency scaling.}
\Description{The impact of CPU and GPU clock frequencies on system performance for ResNet-18. The heatmaps illustrate the non-linear sensitivity of (a) inference latency and (b) power consumption to frequency scaling.}
\label{fig:observation5}
\end{figure}

We conducted an extensive grid search across the CPU and GPU frequency scaling space to characterize their joint impact on inference latency and power consumption. As shown in Figure~\ref{fig:observation5}, inference latency is not uniformly sensitive to frequency scaling. For compute-bound phases, increasing GPU frequency yields significant latency reductions. However, as frequency increases, the system encounters a diminishing returns threshold where memory bandwidth become the primary bottleneck. Specifically, at high GPU frequencies, the performance gain plateaus while power consumption continues to climb quadratically.
We also observe a clear decoupling between minimum latency and minimum energy consumption. While the highest frequency configuration minimizes latency, it often leads to sub-optimal energy efficiency due to the non-linear increase in power. 

These observations reveal the necessity of offline modeler. By mapping specific operator characteristics (such as sparsity and arithmetic intensity) to the relationship between latency and power, SparseDVFS can bypass the inefficiencies of default governors. It allows the system to proactively select the energy-optimal frequency configuration that satisfies the target latency constraint, rather than blindly pursuing maximum performance at the cost of excessive energy waste.

\begin{figure*}[tb]
\large
\centerline{\includegraphics[width=0.85\linewidth]{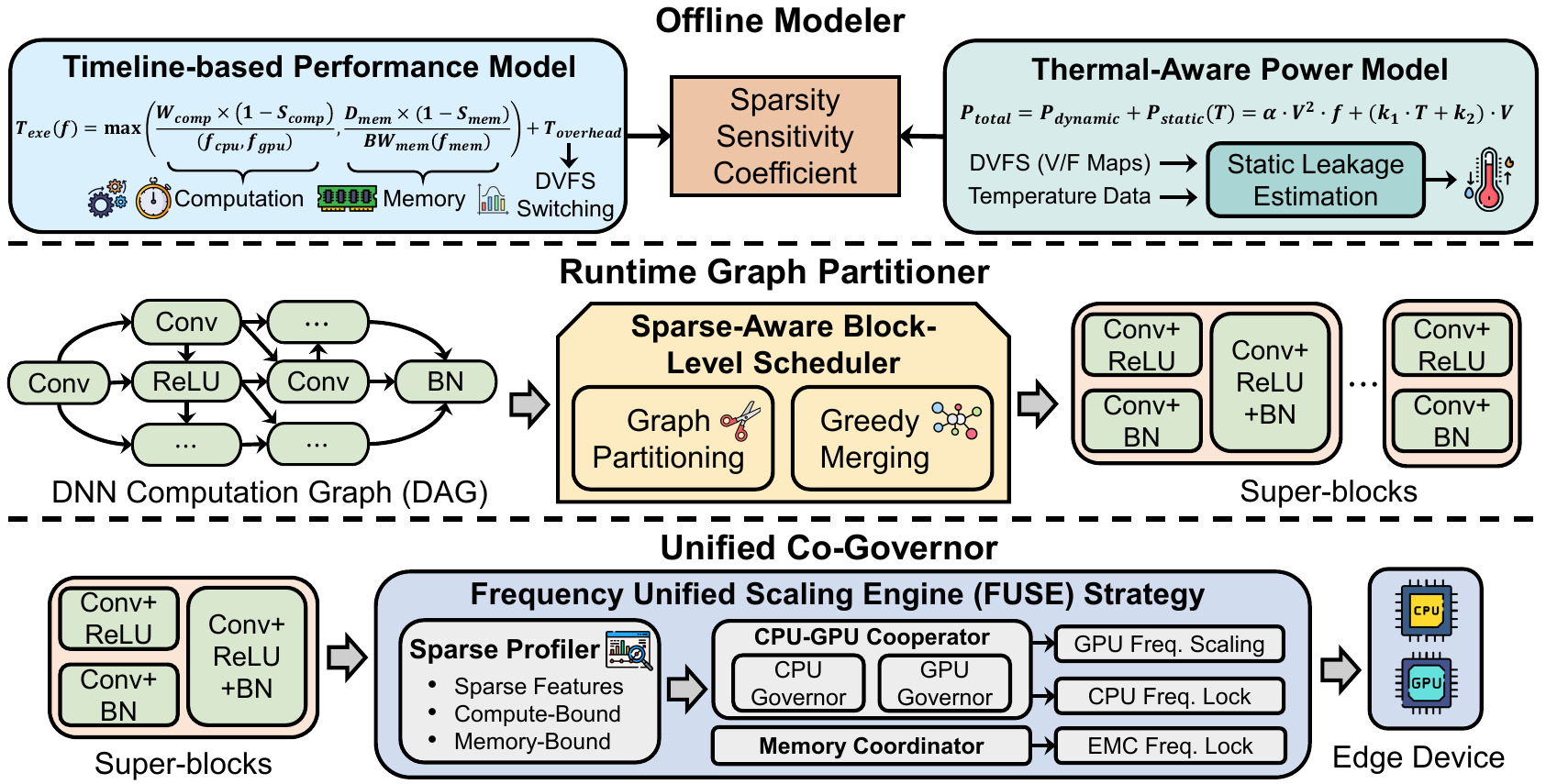}}
\caption{Overview of the SparseDVFS framework. The system is architected in three synergistic stages: (1) the offline modeler establishes a deterministic mapping between operator characteristics and optimal frequency triplets through timeline-based performance analysis and thermal-aware power modeling. (2) the runtime graph partitioner performs greedy operator aggregation to construct super-blocks, ensuring that fine-grained DVFS scaling satisfies the latency amortization constraint. (3) the unified co-governor executes coordinated frequency scaling across heterogeneous components via the frequency unified scaling engine (FUSE) strategy, while masking DVFS switching latency through a pipelined look-ahead mechanism.}
\Description{Overview of the SparseDVFS framework. The system is architected in three synergistic stages: (1) the offline modeler establishes a deterministic mapping between operator characteristics and optimal frequency triplets through timeline-based performance analysis and thermal-aware power modeling. (2) the runtime graph partitioner performs greedy operator aggregation to construct super-blocks, ensuring that fine-grained DVFS scaling satisfies the latency amortization constraint. (3) the unified co-governor executes coordinated frequency scaling across heterogeneous components via the frequency unified scaling engine (FUSE) strategy, while masking DVFS switching latency through a pipelined look-ahead mechanism.}
\label{fig:framework}
\end{figure*}

\section{System Overview} \label{Overview}
SparseDVFS functions as a middleware situated between the deep learning framework (e.g., PyTorch) and the OS interfaces. It is designed to intercept the execution flow of DNN inference, analyze the upcoming workload, and proactively adjust hardware settings before execution begins. Unlike reactive governors that rely on trailing utilization indicators, SparseDVFS uses a predictive model based on static graph analysis and dynamic sparsity profiling.

Figure~\ref{fig:framework} shows the architecture of SparseDVFS, which consists of three components. \ding{182} The offline modeler operates prior to deployment, characterizing the hardware's behavior to build a comprehensive performance and power model. \ding{183} During inference, the runtime graph partitioner dynamically segments the DNN computation graph into super-blocks to amortize switching overheads. \ding{184} Finally, the unified co-governor enforces the optimal frequency configurations for these blocks across the CPU, GPU, and memory subsystems, utilizing a look-ahead mechanism to hide DVFS switching latency. This pipelined design ensures that SparseDVFS can deliver significant energy savings without compromising the strict latency requirements of edge inference.

\section{SparseDVFS Design} \label{Design}
\subsection{Offline Modeler}  \label{sec:Modeler}
This component aims to eliminate the cold-start latency and unpredictability inherent in existing black-box approaches (e.g., deep neural networks)~\cite{kim2021ztt,lin2023workload} by establishing a deterministic, physics-based mapping between operator characteristics and hardware states.

\subsubsection{Timeline-based Performance Model.}
The execution performance of DNN operators on heterogeneous Edge SoCs is fundamentally governed by the interplay between computational throughput and memory bandwidth, a relationship classically described by the Roofline Model. However, traditional Roofline formulations assume dense computation, failing to account for the dynamic workload reduction introduced by algorithmic sparsity. 
To this end, we formulate a timeline-based performance model that explicitly integrates sparsity sensitivity coefficients to predict the execution time $T_{exe}$ of an operator under varying frequency configurations. This model treats execution time not as a monolithic random variable, but as a convex piecewise linear function determined by the bottleneck resource. 

Specifically, the execution time is derived by calculating the maximum latency between the compute-bound execution path and the memory-bound data transfer path, augmented by a system overhead constant.
For a given frequency configuration vector $\mathbf{f} = [f_{cpu},f_{gpu},f_{mem}]$, the predicted execution time is defined as:
\begin{equation}
\footnotesize
T_{exe}(\mathbf{f}) = \max \left( \frac{W_{comp} \cdot (1 - S_{comp})}{\mathcal{P}_{peak}(f_{cpu}, f_{gpu})}, \frac{D_{mem} \cdot (1 - S_{mem})}{\mathcal{B}_{mem}(f_{mem})} \right) + T_{overhead}
\end{equation}
where $W_{comp}$ represents the theoretical computational workload (FLOPs) derived statically from the operator's dimensions, while $S_{comp} \in [0,1]$ is the computational sparsity coefficient, representing the fraction of zero-valued activations that the hardware can effectively skip. $W_{comp} \cdot (1 - S_{comp})$ thus quantifies the effective computational load, revealing the execution slack available for frequency downscaling in sparse layers. 
Conversely, the memory is governed by the total data volume $D_{mem}$ and the storage sparsity coefficient $S_{mem}$. Crucially, $S_{mem}$ captures the bandwidth savings dependent on the sparsity. For structured sparsity, $S_{mem} \approx S_{comp}$, whereas for unstructured sparsity, $S_{mem}$ approaches zero due to random access inefficiencies. 
$\mathcal{P}_{peak}$ and $\mathcal{B}_{mem}$ denote the peak theoretical performance and memory bandwidth at respective frequencies, obtained via offline micro-benchmarking. 
$T_{overhead}$ accounts for constant system latencies such as kernel launch overhead and synchronization barriers.

Table~\ref{tbl:predictor_comparison} shows the performance of our proposed timeline-based performance model. We take the FLOPs-based predictor as the baseline, which utilizes theoretical computational workload to build a simple linear model for latency prediction. As shown, the FLOPs-based predictor only achieves 8.2\%$\sim$18.4\% accuracy across all evaluated models.By integrating sparsity sensitivity and hardware-specific bottlenecks, our deterministic predictor achieves $\pm$10\% accuracies of 86.25\%, 84.6\%, 82.1\%, and 80.8\% for ResNet-18, ResNet-101, ViT-B16, and ViT-L16, respectively. This performance is even comparable with the black-box model nn-Meter, which relies on extensive kernel-level profiling. Note that our physics-based predictor is designed to be extremely lightweight (only 32 bytes), while nn-Meter's size is nearly 856MB which is not suitable for the online prediction on edge devices. 
\begin{table}[htbp]
\footnotesize
\centering
\caption{$\pm 10\%$ latency prediction accuracy and model size of different predictors.}
\label{tbl:predictor_comparison}
\begin{tabular}{@{}cccc@{}}
\toprule[0.75pt]
\textbf{Predictor} & \textbf{Model} & \textbf{$\pm 10\%$ Accuracy} & \textbf{Model Size} \\ 
\midrule[0.6pt]
\multirow{4}{*}{FLOPs-based~\cite{kim2019mulayer}} 
& ResNet-18  & 18.4\%  & \multirow{4}{*}{8B} \\
& ResNet-101 & 12.8\%  &  \\
& ViT-B16    & 11.5\%  &  \\ 
& ViT-L16    & 8.2\%   &  \\ 
\midrule[0.6pt]
\multirow{4}{*}{\begin{tabular}[c]{@{}l@{}}nn-Meter~\cite{zhang2021nn}\end{tabular}} 
& ResNet-18  & 91.5\% & \multirow{4}{*}{\textasciitilde 856MB} \\
& ResNet-101 & 88.1\% &  \\
& ViT-B16    & 87.2\% &  \\ 
& ViT-L16    & 85.9\% &  \\ 
\midrule[0.6pt]
\multirow{4}{*}{Timeline-based (Ours)} 
& ResNet-18  & 86.2\%  & \multirow{4}{*}{32B} \\
& ResNet-101 & 84.6\%  &  \\
& ViT-B16    & 82.1\%  &  \\
& ViT-L16    & 80.8\%  &  \\ 
\bottomrule[0.75pt]
\end{tabular}
\end{table}

\subsubsection{Thermal-Aware Power Model.}
Accurate energy optimization in embedded AI chips requires a precise estimation of power consumption that accounts for the non-linear dependency of static leakage current on temperature. A simplistic model ignoring thermal feedback risks underestimating power draw during intensive inference workloads, leading to thermal throttling and performance jitter. SparseDVFS introduces a thermal-aware power model that decouples total power into dynamic switching power and temperature-dependent static leakage power. The dynamic component is modeled as a function of the activity factor, supply voltage , and frequency , while the static component is modeled as a linear function of temperature  scaled by voltage. The unified power model can be expressed as:
\begin{equation}
\begin{aligned}
P_{total}(f, T) &= P_{dynamic} + P_{static}(T) \\
                &= \alpha(S_{comp}) \cdot V^2 \cdot f + (k_1 \cdot T + k_2) \cdot V
\end{aligned}
\end{equation}
where the activity factor $\alpha$ is explicitly modeled as a function of sparsity $S_{comp}$, reflecting that sparse data streams induce fewer bit-flips in the datapath, thereby reducing dynamic power. 
The coefficients $k_1$ and $k_2$ are device-specific thermal constants derived from regression analysis of offline profiling data, capturing the sensitivity of leakage current to junction temperature and the base leakage at a reference temperature, respectively. 

Our proposed thermal-aware power model enables the system to perform proactive thermal management. When the device temperature $T$ rises, the model predicts the super-linear increase in $P_{static}$, prompting the scheduler to select a voltage-frequency pair that not only satisfies the immediate latency constraint but also minimizes total energy $E = P_{total} \times T_{exe}$ by potentially lowering voltage to reduce the thermal envelope. This feedback loop prevents the device from hitting hardware thermal limits, ensuring sustained real-time performance.

\begin{figure}[tb]
\setlength{\abovecaptionskip}{0pt}
\setlength{\belowcaptionskip}{0pt}
\centering
\subfigure[Inference Latency Prediction]{
\begin{minipage}[b]{0.4825\linewidth}
\includegraphics[width=1\linewidth]{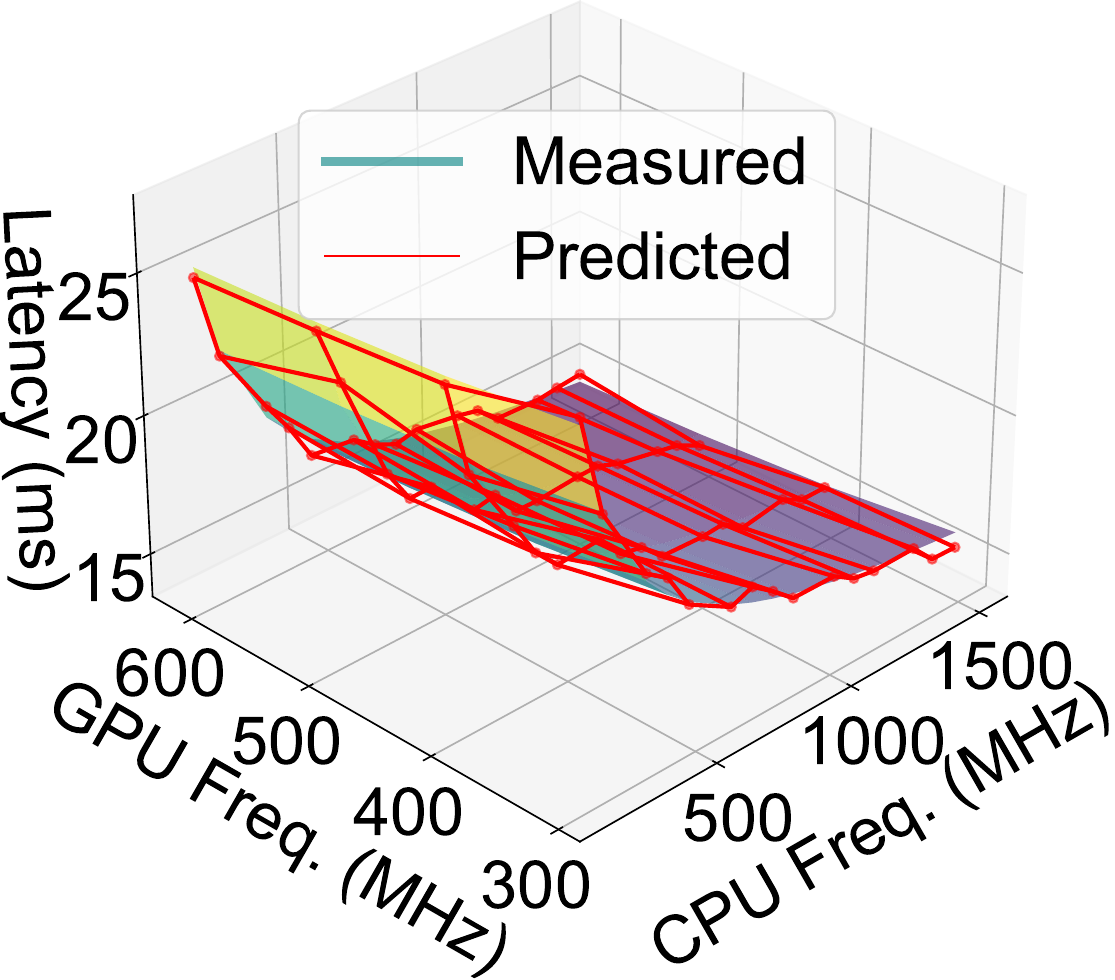}
\end{minipage}}
\subfigure[Thermal-Aware Power Prediction]{
\begin{minipage}[b]{0.4825\linewidth}
\includegraphics[width=1\linewidth]{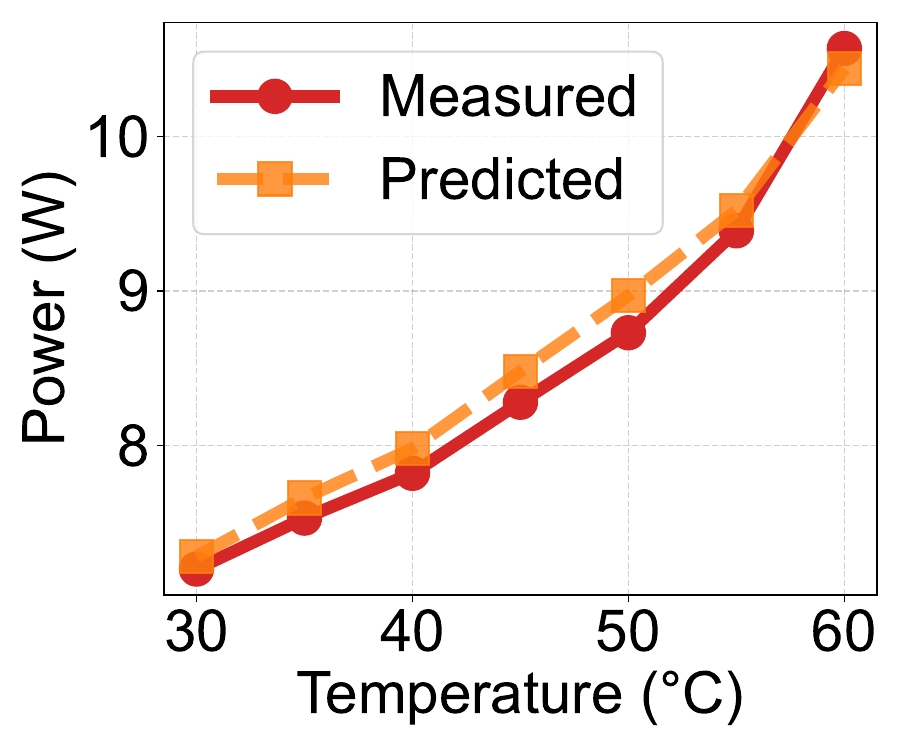}
\end{minipage}}
\caption{An example of offline modeler for ResNet-18. (a) latency prediction under different CPU/GPU frequency combinations and (b) thermal-aware power prediction.}
\Description{An example of offline modeler for ResNet-18. (a) latency prediction under different CPU/GPU frequency combinations and (b) thermal-aware power prediction.}
\label{fig:modeler}
\end{figure}

Figure~\ref{fig:modeler} presents a comparison between measured data and the prediction data of our offline modeler for ResNet-18. As shown in Figure~\ref{fig:modeler}(a), the timeline-based performance model accurately characterizes operator latency across the joint CPU-GPU frequency scaling space,  with a mean error below 2\%. Simultaneously, Figure~\ref{fig:modeler}(b) shows the prediction accuracy of the thermal-aware power model, which successfully captures the super-linear increase in power consumption driven by temperature-dependent static leakage. Overall, our physics-based offline modeler provides the necessary deterministic foundation for proactive DVFS scaling, effectively bypassing the unpredictability and cold-start overheads inherent in traditional black-box learning methods.

\subsection{Runtime Graph Partitioner} \label{sec:Partitioner}
While the offline modeler provides optimal configurations for individual operators, blindly applying these settings at runtime is impractical due to the non-trivial hardware switching latency ($T_{switch}$) associated with voltage regulator adjustments and PLL relocking. To resolve the conflict between fine-grained optimization and switching overhead, the runtime graph partitioner dynamically transforms the fine-grained operator graph into a sequence of execution-friendly super-blocks.

\subsubsection{Greedy Fusion and Super-block Construction.}
The partitioner operates on the DNN directed acyclic graph (DAG) by employing a greedy merging algorithm that traverses the topologically sorted sequence of operators. The core logic dictates that adjacent operators should be fused into a single super-block if they share similar sparsity characteristics, and thus compatible optimal frequencies, or if the execution time of an individual operator is insufficient to amortize the cost of a frequency switch. 

Algorithm \ref{alg:Partitioning} maintains a current block and tentatively adds the next operator in the sequence. For each candidate merge, it evaluates two criteria: homogeneity and amortization. Homogeneity checks if the optimal frequency of the candidate operator, $f_{opt}^{next}$, falls within a tolerance threshold of the current block's frequency $f_{opt}^{curr}$. If the frequencies diverge significantly (e.g., a transition from a sparse, low-frequency ReLU to a dense, high-frequency Convolution), the partitioner considers creating a block boundary. However, this decision is overridden by the Amortization constraint. 
\begin{algorithm}
\caption{Sparse-Aware Greedy Graph Partitioning}
\label{alg:Partitioning}
\footnotesize
\begin{algorithmic}[1]
\REQUIRE $G(V, E)$: DNN computation graph sorted topologically
\REQUIRE $M_{perf}, M_{power}$: Offline performance and power models
\REQUIRE $T_{switch}$: Hardware frequency switching latency
\REQUIRE $N$: Amortization granularity factor
\ENSURE $\mathcal{S}$: Sequence of (super-block, $f_{opt}$) tuples

\STATE $\mathcal{S} \leftarrow \emptyset$, $Block_{curr} \leftarrow \{v_1\}$, $f_{curr} \leftarrow M_{perf}.predict(v_1)$
\FOR{$i = 2$ to $|V|$}
    \STATE $v_{next} \leftarrow V[i]$
    \STATE $f_{next} \leftarrow M_{perf}.predict(v_{next})$
    \STATE $T_{est} \leftarrow M_{perf}.calc\_time(Block_{curr}, f_{curr})$
    
    \STATE \COMMENT{Check amortization and similarity constraints}
    \IF{$T_{est} < N \times T_{switch}$ \OR $f_{next} \approx f_{curr}$}
        \STATE $Block_{curr} \leftarrow Block_{curr} \cup \{v_{next}\}$
        \STATE $f_{curr} \leftarrow \max(f_{curr}, f_{next})$ \COMMENT{Conservative merge}
    \ELSE
        \STATE $\mathcal{S}.append((Block_{curr}, f_{curr}))$
        \STATE $Block_{curr} \leftarrow \{v_{next}\}$
        \STATE $f_{curr} \leftarrow f_{next}$
    \ENDIF
\ENDFOR
\STATE $\mathcal{S}.append((Block_{curr}, f_{curr}))$
\RETURN $\mathcal{S}$
\end{algorithmic}
\end{algorithm}

\subsubsection{Latency Amortization Constraint.}
The partitioner enforces a strict latency amortization constranit to ensure that the energy savings from a DVFS switching latency are not negated by the performance penalty of the switch itself. A super-block is finalized only if its cumulative estimated execution time  satisfies the condition: 
\begin{equation}
T_{block} > N \times T_{switch}
\end{equation}
where $N$ is a tunable granularity factor that dictates the aggressiveness of the aggregation. If the accumulated duration of the current block is less than $N \times T_{switch}$, the partitioner is forced to merge the disparate operator, effectively absorbing the short operator into the long block. 

\begin{figure}[tb]
\large
\centerline{\includegraphics[width=\linewidth]{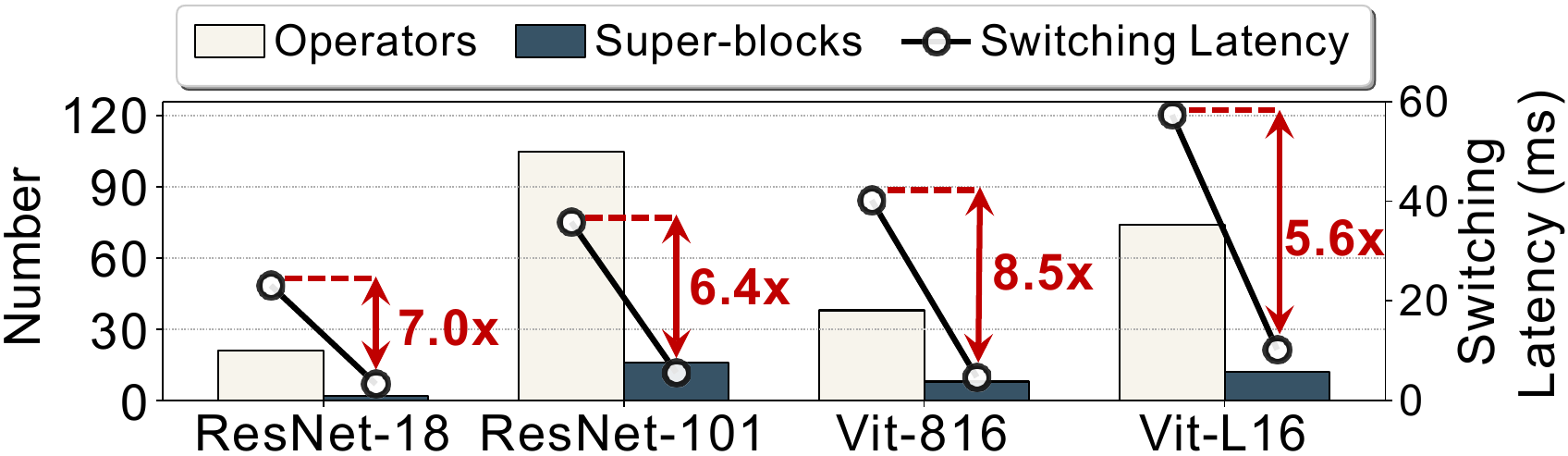}}
\caption{Impact of runtime graph partitioner on operator aggregation granularity and DVFS switching latency. We set aggregation factor $N=5$.}
\Description{Impact of runtime graph partitioner on operator aggregation granularity and DVFS switching latency. We set aggregation factor $N=5$.}
\label{fig:op}
\end{figure}

As shown in Figure~\ref{fig:op}, we analyze the reduction in scaling granularity and the corresponding mitigation of switching overhead across diverse architectures. The partitioner effectively consolidates fine-grained operators into computationally substantial super-blocks to satisfy the latency amortization constraint (aggregation factor $N=5$). Specifically, ResNet-18 is compressed from 21 operators to just 2 super-blocks, while the deeper ResNet-101 is reduced from 105 operators to 16. Similarly, the partitioner aggregates the 38 and 74 operators of ViT-B16 and ViT-L16 into 8 and 12 super-blocks, respectively. 
Crucially, this reduction in granularity translates directly into a drastic decrease in cumulative DVFS switching latency. Compared to operator-level DVFS scaling, SparseDVFS reduces the total switching latency by 7.0$\times$ for ResNet-18 and up to 8.5$\times$ for ViT-B16. These results reveal that by enforcing $T_{block} > N \times T_{switch}$, SparseDVFS successfully amortizes the non-negligible DVFS switching latency.

\subsection{Unified Co-Governor}  \label{sec:Governor}
This component aims to execute the schedule generated by the partitioner. To mitigate the antagonistic effect observed in heterogeneous SoCs (i.e., section~\ref{Antagonistic}), where independent governors for CPU and GPU optimize locally and cause system-level bottlenecks—the governor implements the FUSE (i.e., frequency unified scaling engine) strategy.

The default Linux governors (e.g., schedutil) react to instantaneous load. In a DNN inference pipeline, a GPU-heavy sparse operator might cause CPU utilization to drop, triggering the CPU governor to downclock. When the GPU completes its task and requests new data, the CPU is in a low-power state, causing a wake-up latency spike that stalls the pipeline. Our proposed unified co-governor eliminates this antagonism by treating the frequencies of the CPU, GPU, and Memory as a coupled vector . Instead of reacting to load, it proactively sets the frequency vector based on the known characteristics of the incoming super-block.

\subsubsection{CPU-GPU Cooperative.}
To eliminate antagonistic effects where independent governors create bottlenecks, SparseDVFS implements a proactive race-to-submit policy by treating CPU, GPU, and memory frequencies as a coupled vector. By intercepting CUDA streams and TensorRT execution queues, the co-governor identifies super-block boundaries and immediately locks the CPU to a high-performance state to accelerate prefill tasks like data preprocessing and kernel launches. This prevents the GPU starvation typically caused by reactive governors downclocking the CPU during GPU-intensive phases. Once the kernel is offloaded, the CPU frequency is allowed to drop, effectively masking hardware wake-up latencies and maximizing system-wide energy efficiency without compromising throughput.

\subsubsection{Memory Coordination.}
For sparse operators, the co-governor leverages the insight from the Roofline model that performance is memory-bound.
\begin{itemize}
    \item \textbf{Sparse Blocks:} the governor locks the Memory Controller (EMC) to a high frequency to maximize bandwidth, while simultaneously lowering the GPU core frequency. This configuration saves GPU power without hurting performance, as the GPU was stalling on memory anyway.
    \item \textbf{Dense Blocks:} the governor scales the EMC down to the minimum level required to support the GPU's compute throughput, reducing the significant power draw of the memory interface.
\end{itemize}

\subsubsection{Pipeline-based Look-Ahead Strategy.}
\begin{figure}[tb]
\large
\centerline{\includegraphics[width=\linewidth]{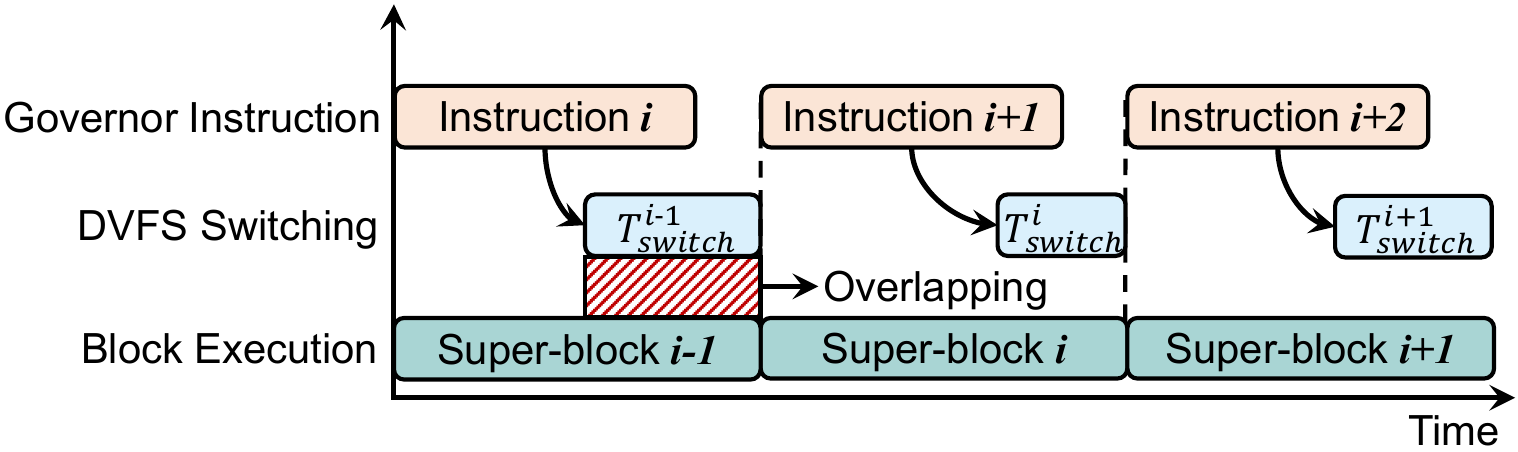}}
\caption{Pipeline-based Look-Ahead DVFS.}
\Description{Pipeline-based Look-Ahead DVFS.}
\label{fig:pipeline}
\end{figure}

To mitigate the impact of DVFS switching latency (e.g., $T_{switch}$), the unified co-governor utilizes a pipeline-based look-ahead mechanism. Traditional DVFS governors typically execute scaling commands serially, causing significant execution stalls as the system waits for voltage and frequency levels to stabilize. As shown in Figure~\ref{fig:pipeline}, SparseDVFS overcomes this bottleneck by employing a pipelined command queue that submits frequency scaling instructions for the subsequent super-block ($i+1$) to hardware registers before the current super-block ($i$) completes its execution. These asynchronous instructions effectively mask the non-negligible DVFS switching latency, without inducing pipeline bubbles. 

The benefits of this strategy are quantified in Table~\ref{tbl:lookahead}. For instance, in ResNet-101, traditional serial DVFS switching incurs a cumulative latency of 10.81ms, which SparseDVFS drastically reduces to just 1.45ms. Similarly, for ViT-L16, the switching overhead is reduced from 7.92ms to 1.08ms. By overlapping the physical switching period with active computation, SparseDVFS ensures that the energy efficiency gains achieved through fine-grained frequency modulation are not negated by system-level performance penalties.

\begin{table}[t]
\centering
\small
\caption{Comparison of DVFS Switching Latency between Traditional DVFS (serial execution) and Pipeline-based Look-Ahead. We set aggregation factor $N=5$.}
\label{tbl:lookahead}
\begin{tabular}{ccc}
\toprule[0.75pt]
\textbf{Model} & \textbf{Traditional DVFS} & \textbf{Pipeline Look-Ahead} \\
\midrule[0.6pt]
ResNet-18   & 7.23ms   & \textbf{0.12ms} \\
ResNet-101  & 10.81ms  & \textbf{1.45ms} \\
ViT-B16     & 5.44ms   & \textbf{0.72ms} \\
ViT-L16     & 7.92ms   & \textbf{1.08ms} \\
\bottomrule[0.75pt]
\end{tabular}
\end{table}

\section{Implementation} \label{sec:Implementation}
SparseDVFS is implemented as a lightweight middleware situated between the deep learning framework and the Linux kernel frequency governors. 
The offline modeler and the runtime graph partitioner are implemented in Python, leveraging the PyTorch 2.1 for computation graph extraction and execution time prediction.
The unified co-governor is implemented in C++ to minimize overhead during inference.
The system is deployed on an NVIDIA Jetson Orin Nano edge device (JetPack 6.0 with Linux Kernel 5.15). This edge platform features an Ampere architecture GPU and an ARM Cortex CPU, sharing unified 8GB DRAM. 

Unlike existing DVFS governors that operate entirely within the OS kernel space, SparseDVFS operates in userspace by intercepting execution streams. Specifically, we detect upcoming operator sequences by intercepting CUDA streams~\cite{nvidia_cuda} and TensorRT execution queues~\cite{nvidia_tensorrt}. 
During runtime, the unified co-governor applies the FUSE strategy by directly manipulating the device's frequency scaling interfaces. We utilized the \textit{jetson\_stats}~\cite{bonghi_jetson_stats}, specifically \textit{jtop.power()} API, to monitor real-time power consumption across the CPU, GPU, and memory rails. 
To bypass the default governors (such as \textit{Linux schedutil}~\cite{gouicem2020fewer} for the CPU governor and \textit{simple\_ondemand}~\cite{pallipadi2006ondemand} for the GPU governor), SparseDVFS takes direct control of the clock frequencies via the \textit{jetson\_clocks} provided by NVIDIA \textit{nvpmodel}~\cite{nvidia_nvpmodel_guide} and \textit{sysfs} nodes. 

\section{Performance Evaluation} \label{sec:Evaluation}
\subsection{Experiment Setup}
\textbf{Models and Datasets.}
In Table~\ref{tbl:model}, our evaluation use CNN-based (e.g., ResNet-18 and ResNet-101~\cite{he2016deep}) and Transformer-based (e.g., ViT-B16 and ViT-L16~\cite{dosovitskiy2020image}) architectures to show the adaptability of SparseDVFS to different sparsities.
All models are pre-trained on ImageNet-2012~\cite{krizhevsky2012imagenet} and converted to ONNX~\cite{onnx_project} before being optimized via the SparseDVFS.
\begin{table}[t]
\footnotesize
\centering
\caption{Detailed configuration of four DNN models.}
\begin{tabular}{cccc} 
\toprule[0.75pt] 
\textbf{Model} & \textbf{Op. Count} & \textbf{FLOPs (G)} & \textbf{Size (M)} \\ 
\midrule[0.6pt]
ResNet-18   & 21   & 1.82    & 11.69   \\
ResNet-101  & 105  & 7.87    & 44.55  \\
ViT-B16     & 38   & 11.29   & 58.07   \\
ViT-L16     & 74   & 39.86   & 203.36  \\ 
\bottomrule[0.75pt]
\end{tabular}
\label{tbl:model}
\end{table}


\fakepar{Frequency Configurations}
To comprehensively explore the optimization space, we utilize the full range of accessible frequencies on the Jetson Orin Nano. 
For CPU governor, there are 20 distinct frequency levels ranging from 115MHz to 1510MHz. For GPU governor, there 5 distinct frequency levels: 306/408/510/612/624MHz.

\fakepar{Baselines}
We compare SparseDVFS against four power management strategies:
\begin{itemize}
    \item \textbf{Default DVFS:} The standard \textit{Linux schedutil} governor for the CPU and NVIDIA's \textit{simple\_ondemand} governor for the GPU, which is our default baseline.
    \item \textbf{nvpmodel (MAX-N)~\cite{nvidia_nvpmodel_guide}:} A static maximum performance mode provided by NVIDIA JetPack that locks all clocks to their highest frequencies.
    \item \textbf{GearDVFS~\cite{lin2023workload}:} A deep reinforcement learning (DRL)-based governor optimized for concurrent workloads that operates at a coarse model-level granularity.
    \item \textbf{Ascend-DVFS~\cite{wang2025using}:} An state-of-the-art fine-grained operator-level DVFS governor employing a genetic algorithm for frequency scaling.
\end{itemize}

\subsection{Comparison of End-to-End Latency}
\begin{figure}[tb]
\large
\centerline{\includegraphics[width=\linewidth]{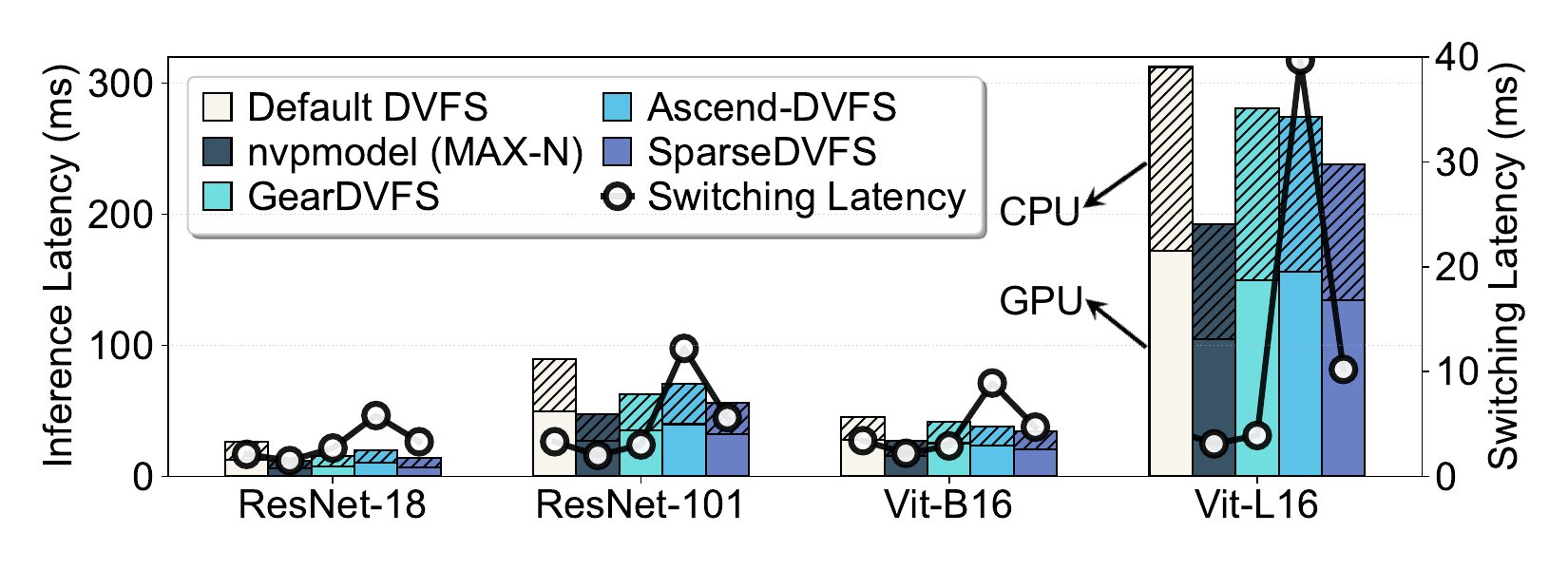}}
\caption{Comparison of end-to-end inference latency across different DNN models.}
\Description{Comparison of end-to-end inference latency across different DNN models.}
\label{fig:latency}
\end{figure}

As shown in Figure~\ref{fig:latency}, SparseDVFS demonstrates the ability to maintain competitive latency while aggressively saving power. Specifically, nvpmodel (MAX-N) achieved the lowest latency by running all clocks at maximum. SparseDVFS incurred a modest latency penalty of approximately 12.8\% compared to MAX-N.
However, it consistently outperformed the Default DVFS and matched the latency profile of GearDVFS. This slight increase is a calculated trade-off, SparseDVFS intentionally slows down sparse blocks where the "race-to-idle" strategy of MAX-N yields diminishing returns.

\subsection{Comparison of Energy and Power}
\begin{figure}[tb]
\large
\centerline{\includegraphics[width=\linewidth]{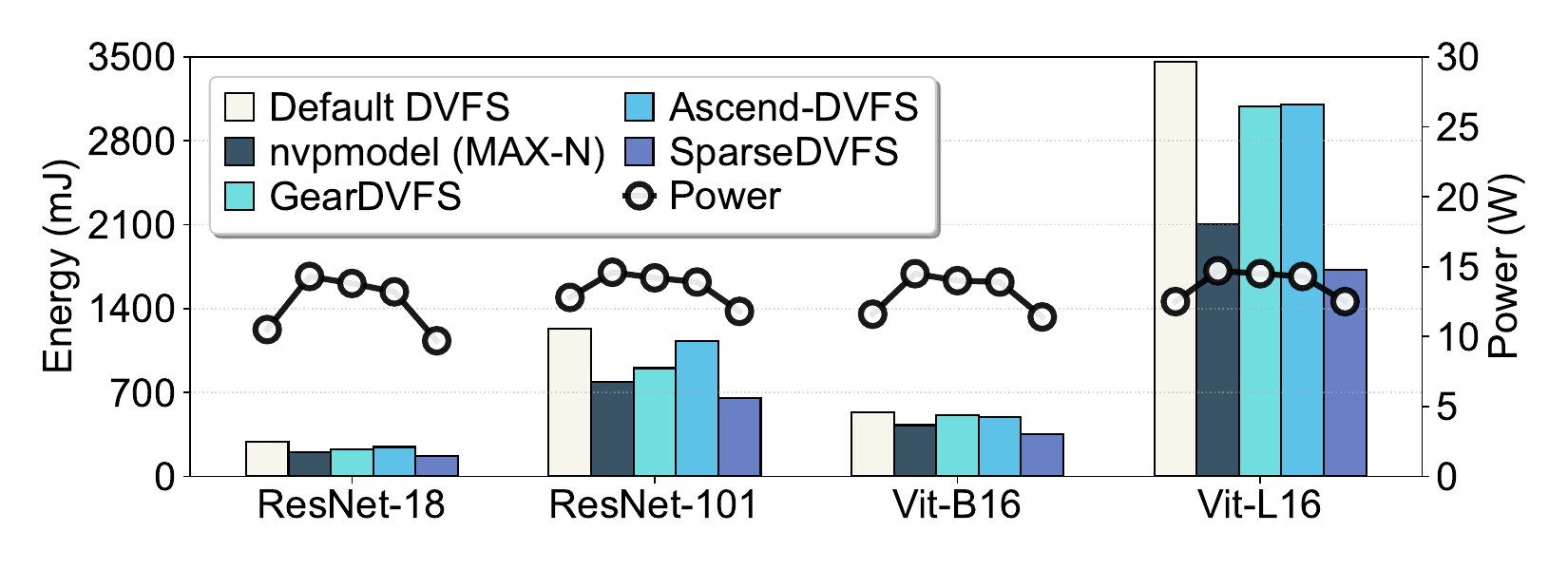}}
\caption{Energy and power profile comparison across different DNN models.}
\Description{Energy and power profile comparison across different DNN models.}
\label{fig:energy_power}
\end{figure}

As shown in Figure~\ref{fig:energy_power}, Energy consumption is where SparseDVFS exhibits its dominance. The power profile of SparseDVFS is significantly lower and more stable than the baselines. While nvpmodel consistently draws peak power (~15W) and GearDVFS fluctuates in the 10-12W range, SparseDVFS dynamically modulates power between 7W and 10W depending on the operator sparsity. By reducing frequency during high-sparsity phases (e.g., ReLU-heavy blocks), it drastically cuts the integrated energy area under the curve.

\subsection{Energy Efficiency Gain}
\begin{figure}[tb]
\large
\centerline{\includegraphics[width=\linewidth]{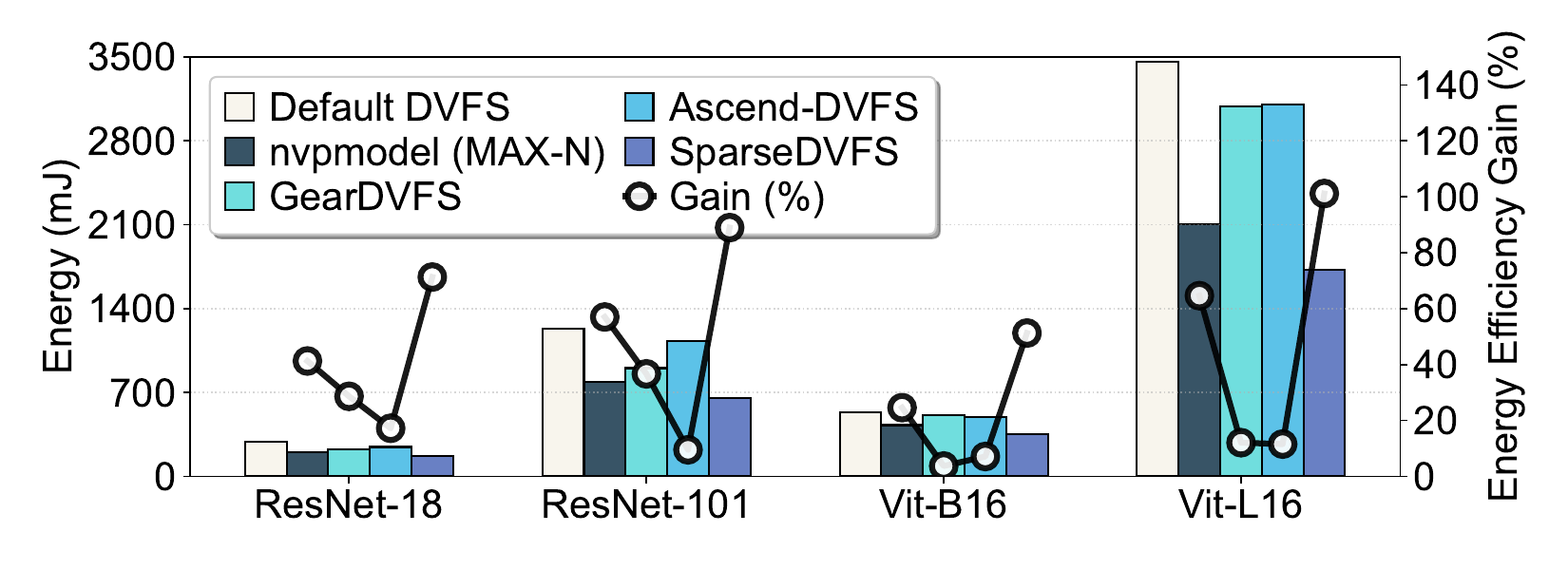}}
\caption{Normalized energy efficiency gain relative to the Default DVFS baseline.}
\Description{Normalized energy efficiency gain relative to the Default DVFS baseline.}
\label{fig:efficient_gain}
\end{figure}

In this section, We quantify energy efficiency gain as the percentage reduction in total energy consumption relative to the Default DVFS baseline. As shown in Figure~\ref{fig:efficient_gain}, SparseDVFS achieves an impressive average energy efficiency gain of 78.17\%. In comparison, nvpmodel achieves a 46.86\% gain (mostly due to faster execution time), GearDVFS achieves 20.25\%, and Ascend-DVFS achieves 11.33\%. This substantial margin highlights the efficacy of the sparse-aware block-level approach over coarser model-level strategies.

\subsection{Cost-Gain Analysis}
\begin{figure}[tb]
\large
\centerline{\includegraphics[width=\linewidth]{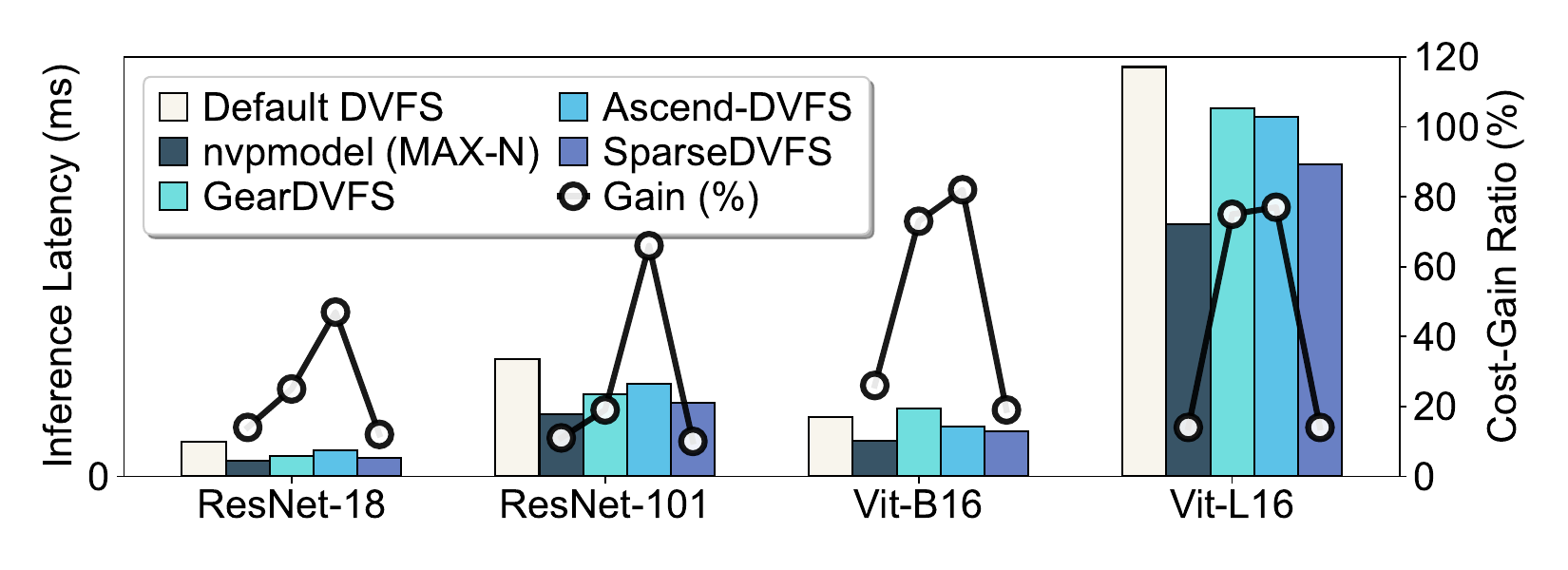}}
\caption{Comparison of Cost-Gain Ratio across different DVFS strategies.}
\Description{Comparison of Cost-Gain Ratio across different DVFS strategies.}
\label{fig:cost_gain_analysis}
\end{figure}

To evaluate the trade-off between performance and efficiency, we utilize the Cost-Gain Ratio~\cite{davaslioglu2014quantifying} (lower is better), which measures the latency cost incurred for every unit of energy gain. Figure~\ref{fig:cost_gain_analysis}, SparseDVFS achieves a superior Cost-Gain Ratio of 14\%. In contrast, GearDVFS (48\%) and Ascend-DVFS (68\%) incur significantly higher latency penalties for their respective energy savings. nvpmodel achieves 16\%, close to SparseDVFS, but requires running at thermal limits.

\subsection{Ablation Study}
\begin{figure}[tb]
\setlength{\abovecaptionskip}{0pt}
\setlength{\belowcaptionskip}{0pt}
\centering
\subfigure[DOTA-v1.0 Dataset]{
\begin{minipage}[b]{0.485\linewidth}
\includegraphics[width=1\linewidth]{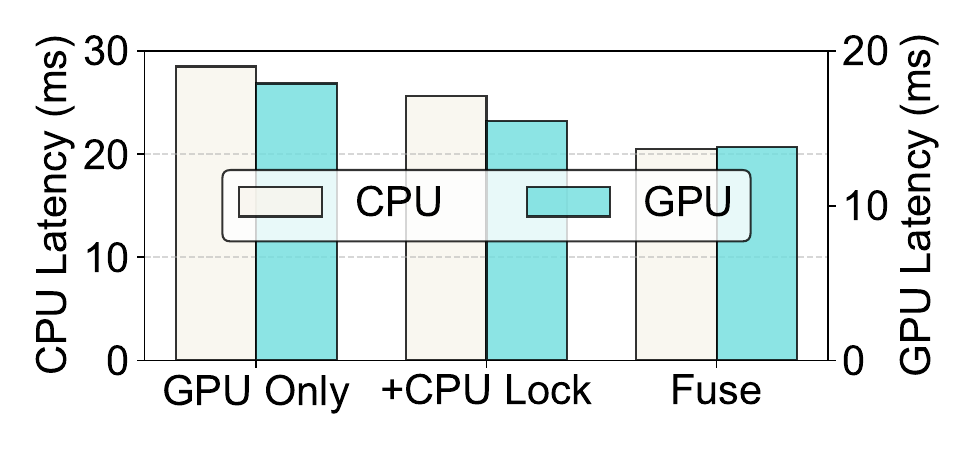}
\end{minipage}}
\subfigure[VisDrone 2019 Dataset]{
\begin{minipage}[b]{0.485\linewidth}
\includegraphics[width=1\linewidth]{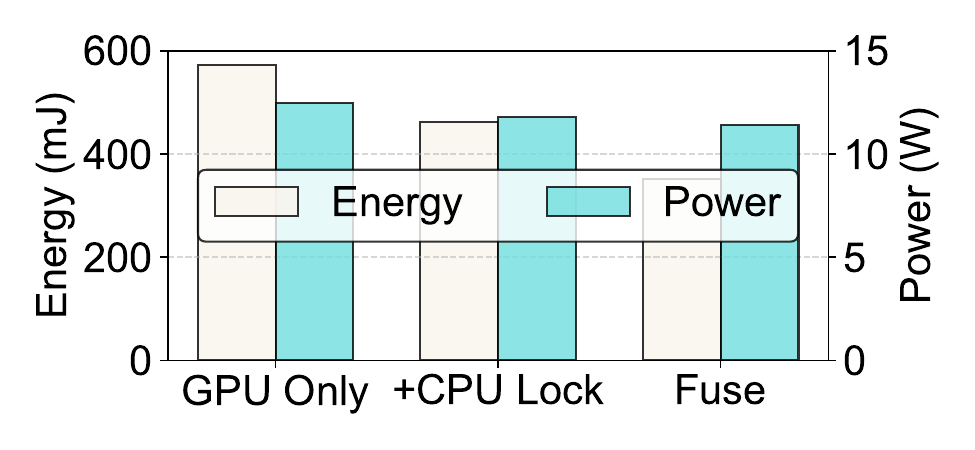}
\end{minipage}}
\caption{Ablation studies of each component of the unified co-governor on (a) DOTA-v1.0 and (b) VisDrone 2019 datasets, respectively. We use ViT-B16 as test load.}
\Description{Ablation studies of each component of the unified co-governor on (a) DOTA-v1.0 and (b) VisDrone 2019 datasets, respectively. We use ViT-B16 as test load.}
\label{fig:ablation}
\end{figure}

In this section, we conducted an ablation study on ViT-B16 using the DOTA-v1.0 and VisDrone 2019 datasets to validate the unified co-governor components. As shown in Figure~\ref{fig:ablation}, While \textit{GPU Only} scaling suffers from the CPU-GPU antagonistic effect, where independent CPU frequency drops during GPU-intensive phases induce wake-up latency spikes, the \textit{+CPU Lock} configuration mitigates GPU starvation by maintaining the CPU in a high-performance state during critical kernel launch windows. Ultimately, the full Fuse (FUSE) strategy, which treats CPU, GPU, and memory frequencies as a coupled vector, achieves optimal energy efficiency by synchronizing transitions and minimizing total integrated power consumption across diverse edge workloads.

\subsection{Thermal Throttling for Stability Analysis}
\begin{figure}[tb]
\large
\centerline{\includegraphics[width=\linewidth]{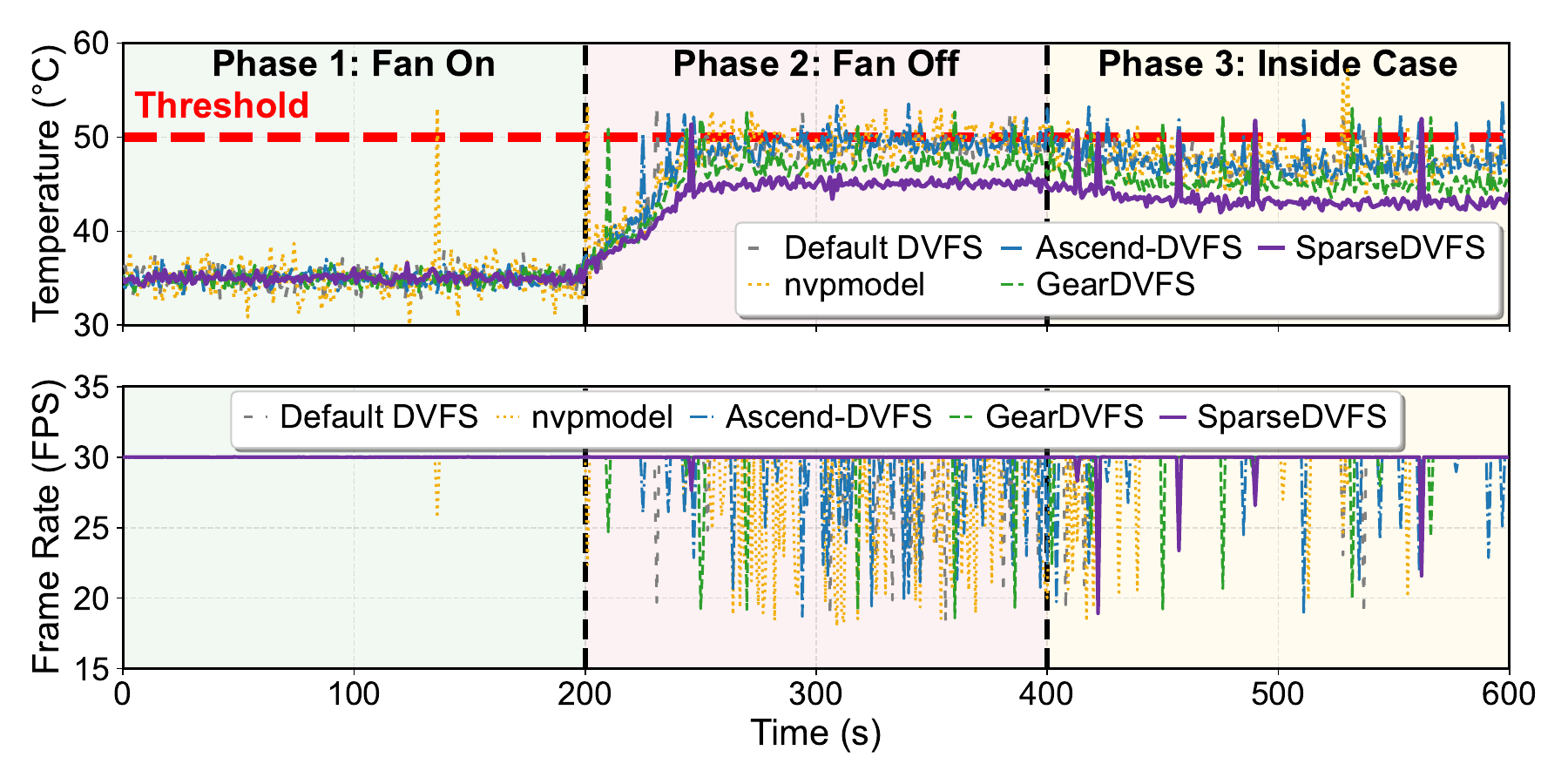}}
\caption{Thermal stability and frame rate jitter analysis of NVIDIA Jetson Orin Nano under sustained load (ViT-B16).}
\Description{Thermal stability and frame rate jitter analysis under sustained load (ViT-B16).}
\label{fig:Thermal_Stability}
\end{figure}

We evaluate the thermal throttling of SparseDVFS for stability analysis. As shown in Figure~\ref{fig:Thermal_Stability}, the nvpmodel and Default DVFS quickly reached the thermal throttle temperature, leading to severe frame rate jitter. SparseDVFS, due to its lower average power draw, maintained a stable temperature trajectory, delaying the onset of thermal throttling significantly. The jitter level for SparseDVFS was the lowest among all methods, proving its suitability for passively cooled edge devices.

\subsection{Impact of Operator Aggregation Granularity}
\subsubsection{Number of Super-blocks.}
\begin{figure}[tb]
\large
\centerline{\includegraphics[width=0.75\linewidth]{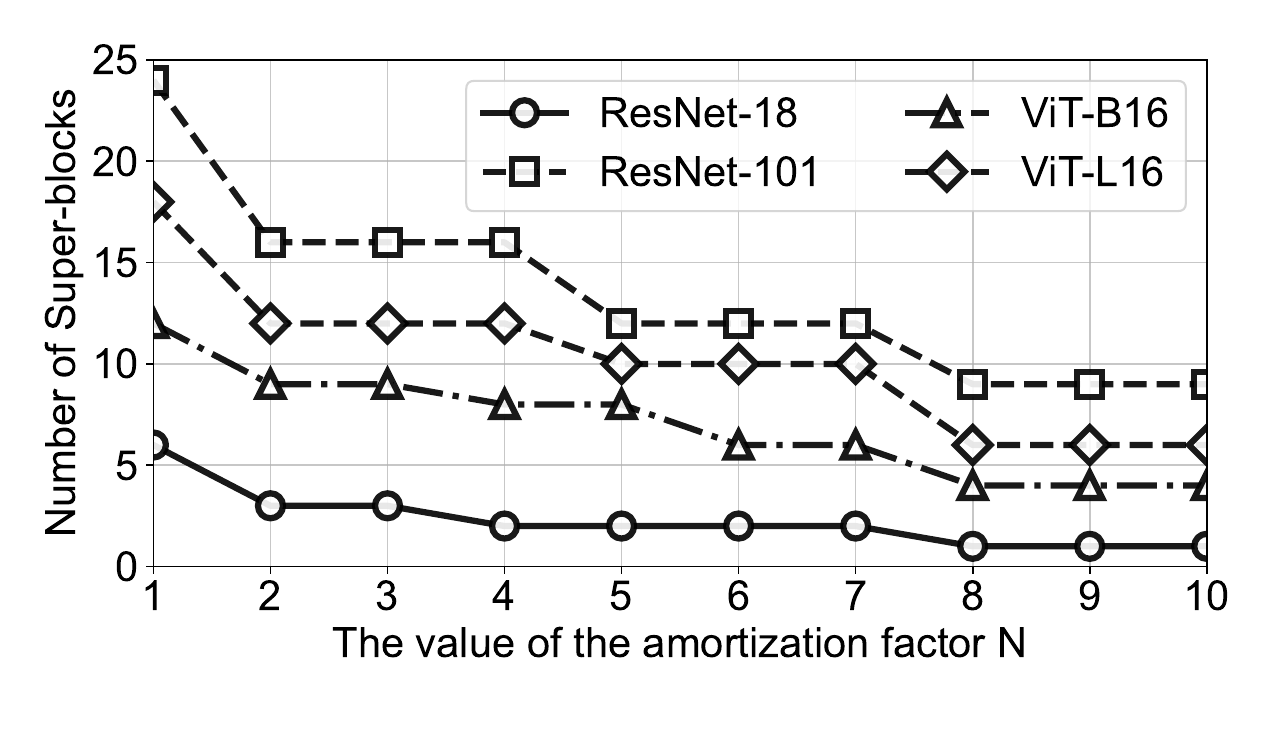}}
\caption{Impact of the amortization factor $N$ on super-block aggregation granularity. Increasing $N$ dictates the trade-off between operator aggregation granularity and DVFS switching latency.}
\Description{Impact of the amortization factor $N$ on super-block aggregation granularity. Increasing $N$ dictates the trade-off between operator aggregation granularity and DVFS switching latency.}
\label{fig:N_gooups}
\end{figure}

The amortization factor $N$ regulates the trade-off between scaling granularity and DVFS switching latency. Figure~\ref{fig:N_gooups} shows a consistent inverse correlation between $N$ and super-block count. For layer-dense models like ResNet-101, small $N$ (e.g., $N=1$) captures fine-grained sparsity but incurs excessive switching risks, increasing $N$ to 10 effectively consolidates operators into computationally significant units. Conversely, ViT-L16 exhibits a more gradual aggregation curve. Because its Transformer blocks possess substantial execution times, the amortization constraint (i.e., $T_{block} > N \times T_{switch}$) is easily satisfied even at lower $N$ values.

\subsubsection{Inference Latency and Switching Overhead.}
\begin{figure}
\setlength{\abovecaptionskip}{0pt}
\setlength{\belowcaptionskip}{0pt}
\centering
\subfigure[ResNet-18]{
\begin{minipage}[b]{0.485\linewidth}
\includegraphics[width=1\linewidth]{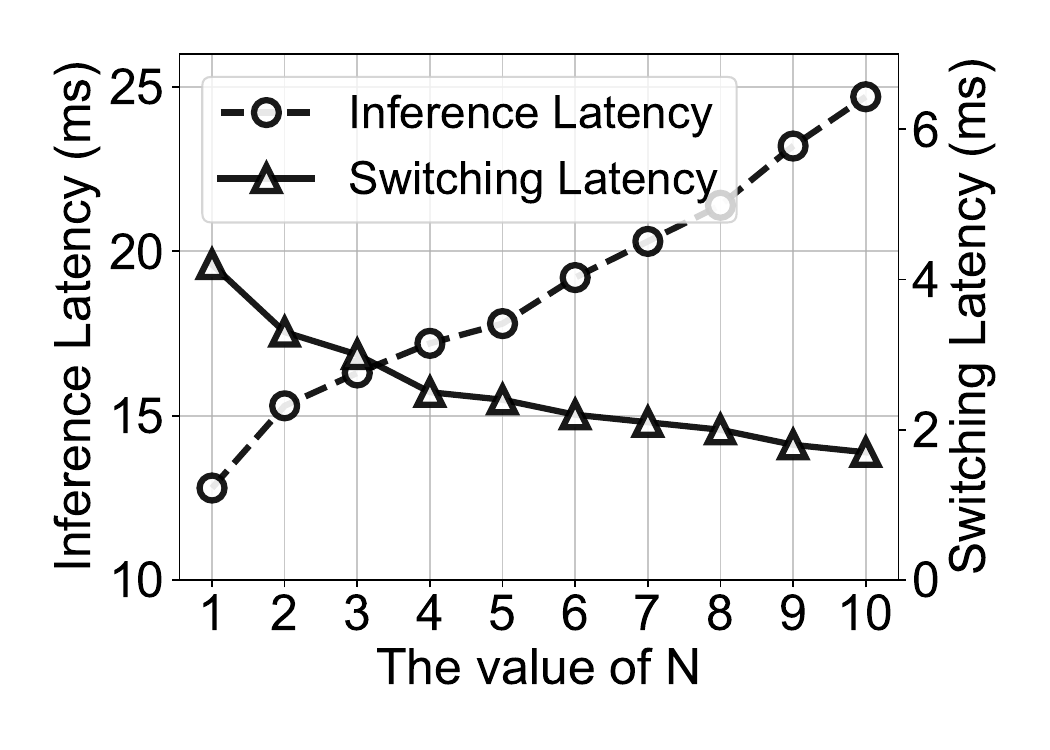}
\end{minipage}}
\subfigure[ResNet-101]{
\begin{minipage}[b]{0.485\linewidth}
\includegraphics[width=1\linewidth]{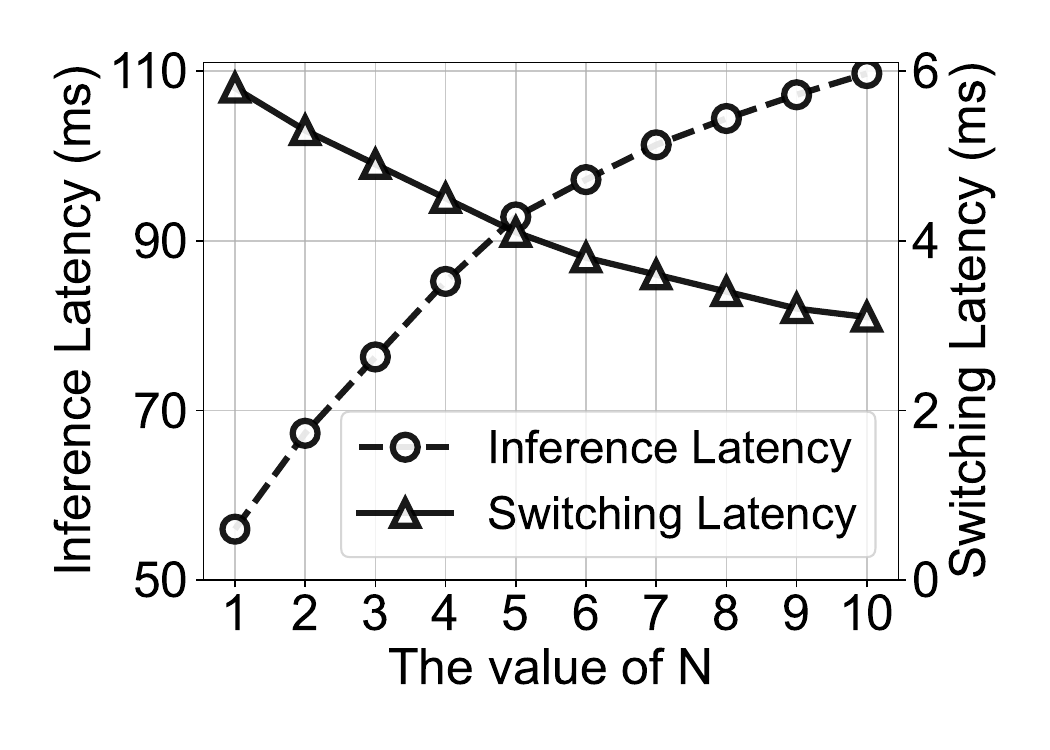}
\end{minipage}}
\subfigure[ViT-B16]{
\begin{minipage}[b]{0.485\linewidth}
\includegraphics[width=1\linewidth]{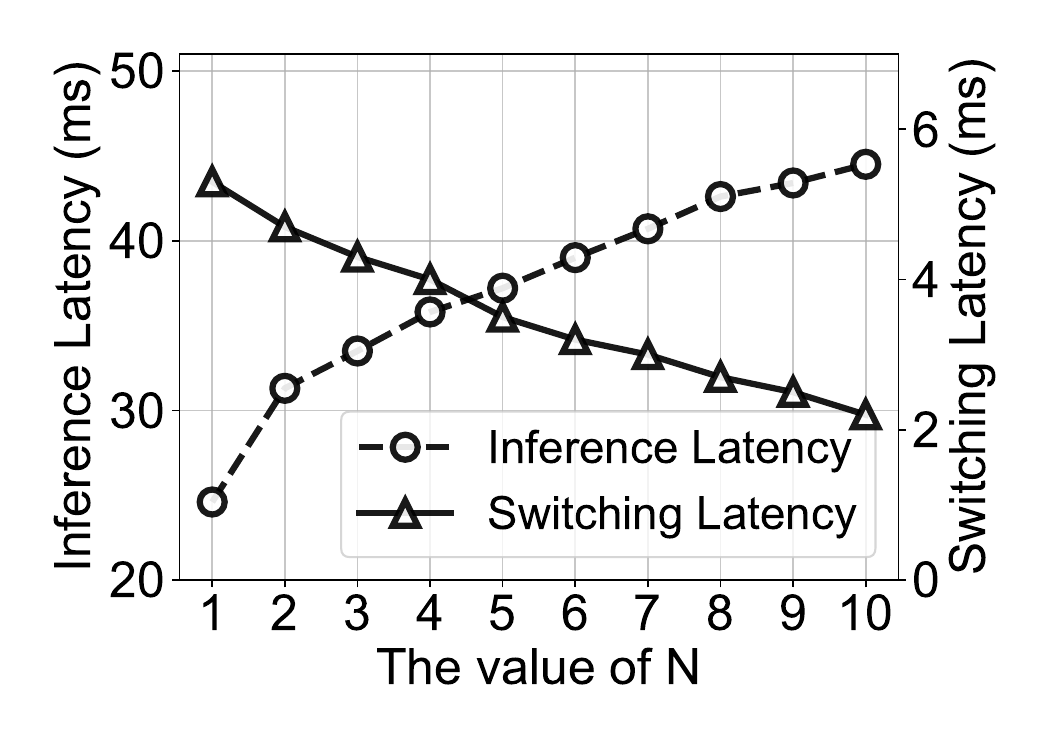}
\end{minipage}}
\subfigure[ViT-L16]{
\begin{minipage}[b]{0.485\linewidth}
\includegraphics[width=1\linewidth]{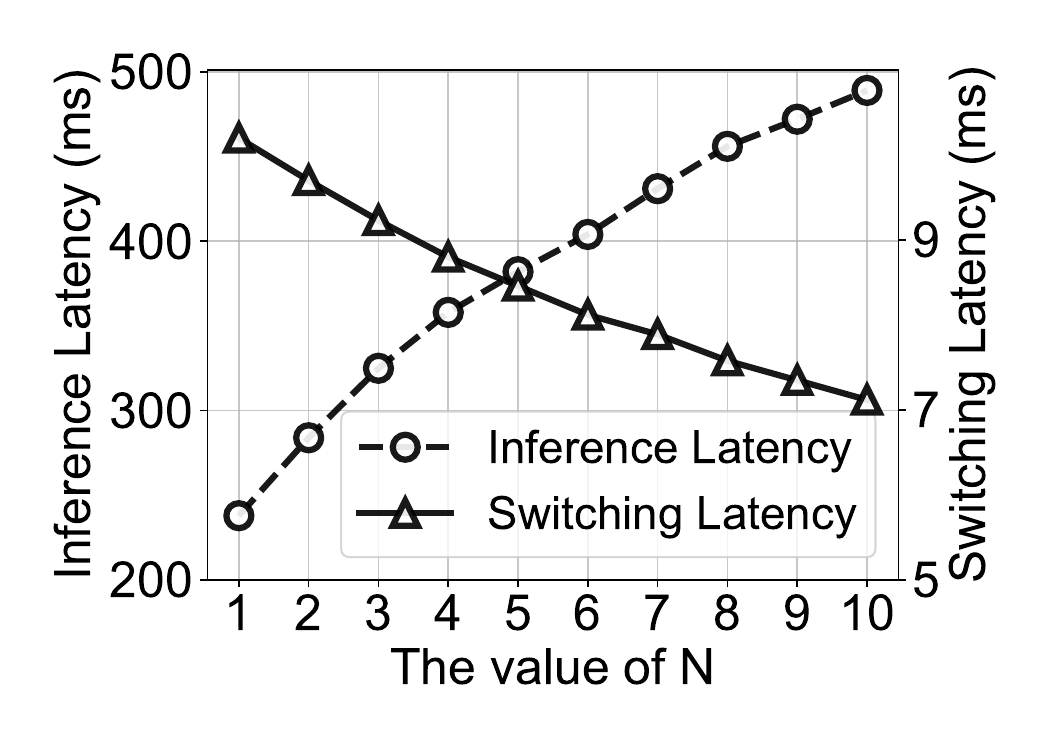}
\end{minipage}}
\caption{Trade-off between inference latency and switching overhead relative to $N$.}
\Description{Trade-off between inference latency and switching overhead relative to $N$.}
\label{fig:N_latency}
\end{figure}

Figure~\ref{fig:N_latency} illustrates how $N$ balances inference latency and switching overhead. At $N=1$, switching latency dominates the timeline, particularly for ResNet-101. 
Although ViT-L16’s bigger operators buffer against switching costs, the Look-Ahead mechanism remains essential for masking transitions between its massive MLP and Attention blocks. Notably, an excessively high $N$ induces a frequency lag effect, where the framework misses downclocking opportunities by merging distinct and dense phases into a single block.

\subsubsection{Power and Energy.}
\begin{figure}
\setlength{\abovecaptionskip}{0pt}
\setlength{\belowcaptionskip}{0pt}
\centering
\subfigure[ResNet-18]{
\begin{minipage}[b]{0.485\linewidth}
\includegraphics[width=1\linewidth]{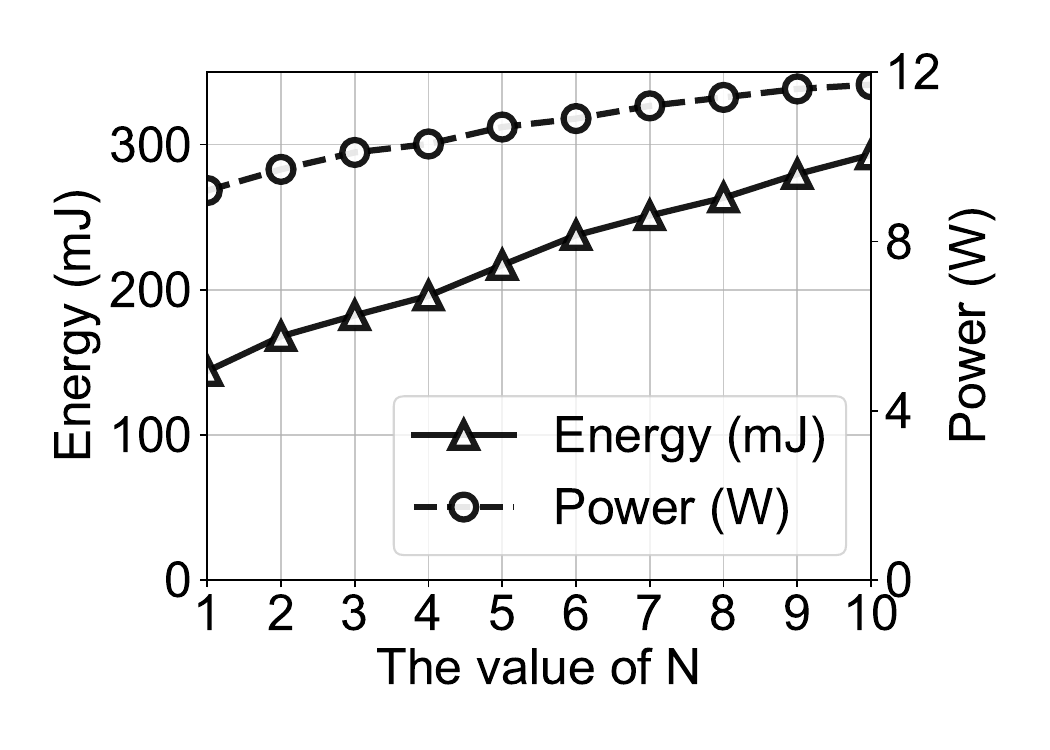}
\end{minipage}}
\subfigure[ResNet-101]{
\begin{minipage}[b]{0.485\linewidth}
\includegraphics[width=1\linewidth]{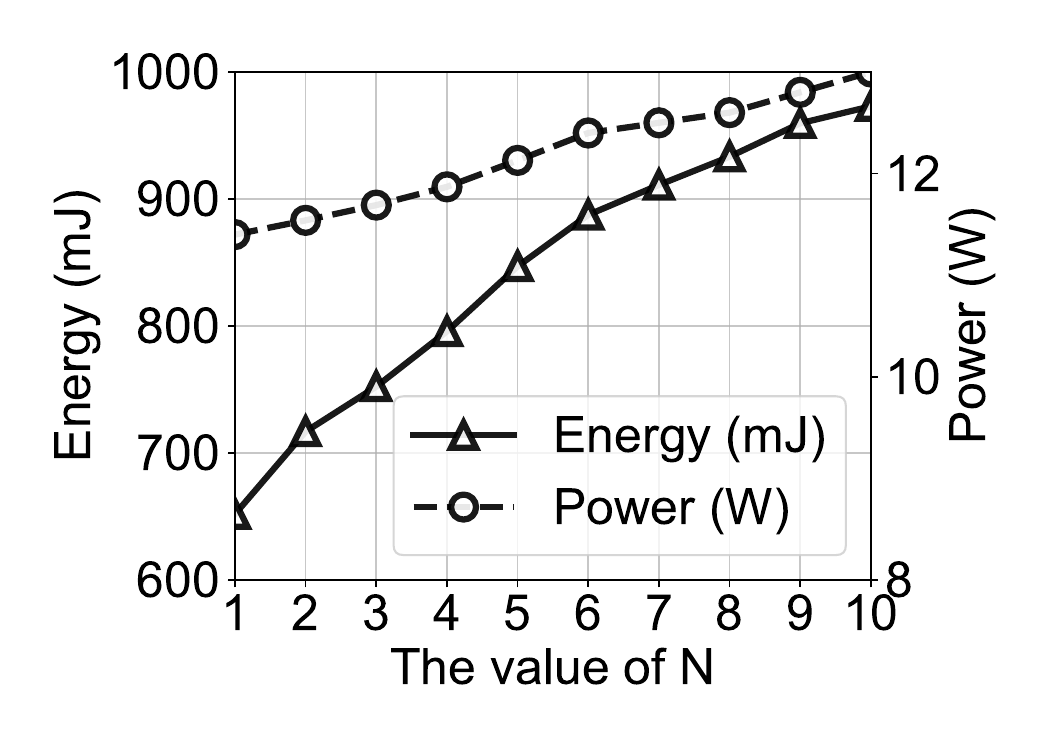}
\end{minipage}}
\subfigure[ViT-B16]{
\begin{minipage}[b]{0.485\linewidth}
\includegraphics[width=1\linewidth]{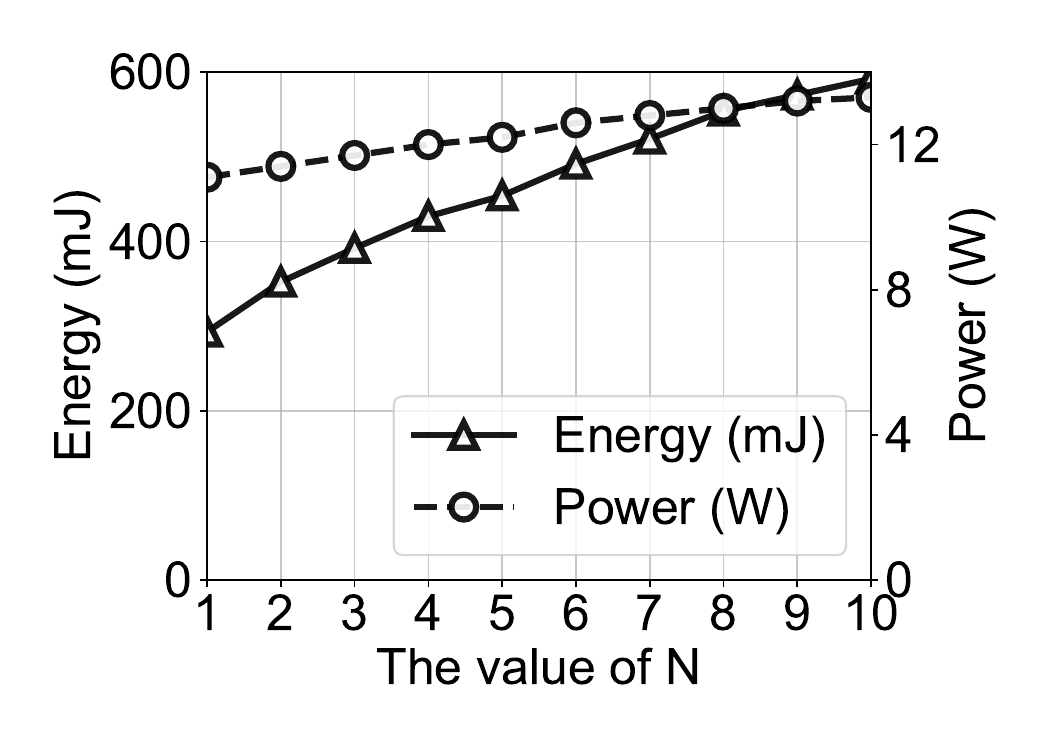}
\end{minipage}}
\subfigure[ViT-L16]{
\begin{minipage}[b]{0.485\linewidth}
\includegraphics[width=1\linewidth]{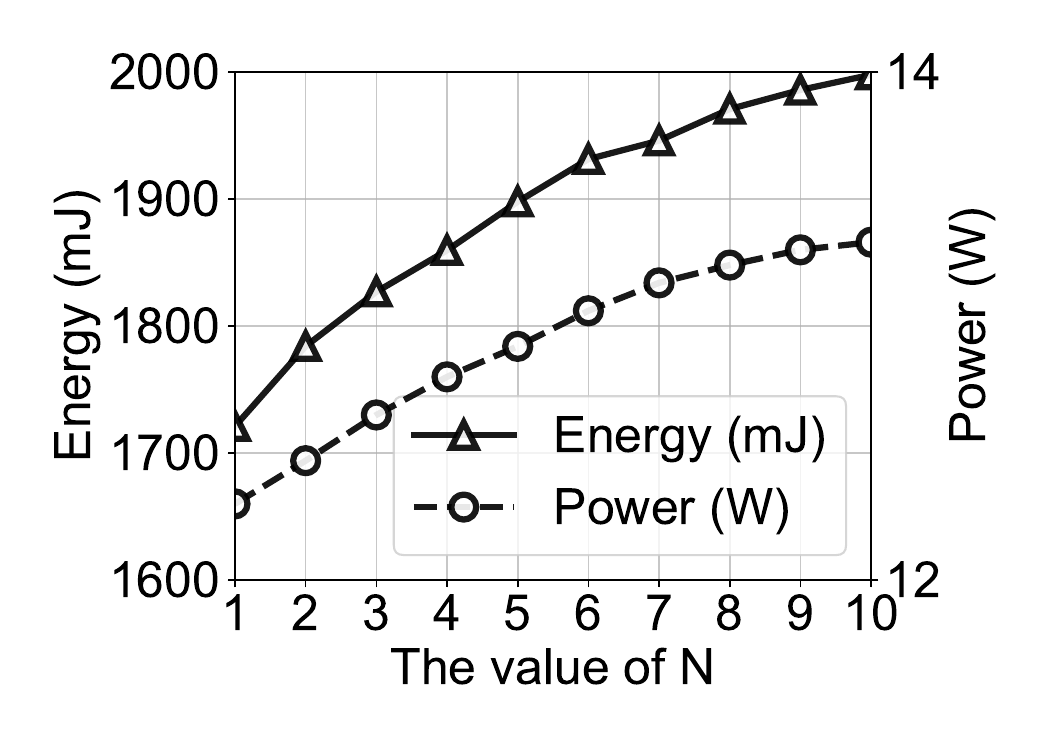}
\end{minipage}}
\caption{Impact of the aggregation factor $N$ on system power and energy consumption.}
\Description{Impact of the aggregation factor $N$ on system power and energy consumption.}
\label{fig:N_power_energy}
\end{figure}

As shown in Figure~\ref{fig:N_power_energy}, the energy-per-inference displays a characteristic U-shaped curve relative to $N$. At low $N$ values, the system suffers from high energy consumed during the frequent high-voltage transitions between blocks. Conversely, at very high $N$ values, the lack of granularity prevents the unified co-governor from identifying short-lived sparse windows, forcing the GPU to remain in high-power states even during memory-bound phases. For the computationally heavy ViT-L16, the energy savings are most pronounced at moderate $N$ settings.

\section{Related work} \label{sec:Related}
\textbf{Performance Modeling for Edge Inference.}
Existing studies like nn-Meter~\cite{zhang2021nn} and nn-stretch~\cite{wei2023nn} rely on kernel-level profiling, while EdgeFM~\cite{yang2023edgefm}, lm-Meter~\cite{wang2025lm} and nnPerf~\cite{chu2023nnperf} focus on macro-level latency. Unlike black-box models or memory-centric simulations like Symphony~\cite{pellauer2023symphony} and Spatula~\cite{feldmann2023spatula}, SparseDVFS implements a white-box and sparsity-aware modeler. It maps operator-level sparsity to optimal frequency triplets with deterministic and lightweight lookup complexity suited for real-time edge execution.

\fakepar{DVFS Technology for AI Accelerators}
Frequency scaling has evolved from coarse model-level adaptation (e.g., zTT~\cite{kim2021ztt} and GearDVFS~\cite{lin2023workload}) to prohibitive layer- or operator-level scaling (e.g., Ascend-DVFS~\cite{wang2025using} and CLONE~\cite{tian2025clone}). To eliminate component interference, FUSE~\cite{zhang2025dissecting} and collaborative frameworks like DVFO~\cite{zhang2024dvfo} and E4~\cite{zhang2025e4} have been proposed.
While prior arts focus on high-overhead operator-level scaling or low-precision model-level governors, SparseDVFS aggregates operators into super-blocks and employing a look-ahead mechanism, it effectively amortizes the non-negligible DVFS switching latency.

\fakepar{Sparse Optimization for DNN Models}
Exploiting model sparsity and low-precision representation is a primary lever for reducing computational and memory intensity of DNN models. 
Algorithmic efforts like nmSPARSE~\cite{lin2023efficient}, Sparse VILA~\cite{khaki2025sparsevila}, and FlashAttention-2~\cite{dao2023flashattention} focus on designing specialized sparse kernels or hardware primitives like SparseCore~\cite{jouppi2023tpu}. 
While these works focus on designing specialized sparse kernels, SparseDVFS treats operator sparsity as a dynamic control signal for DVFS. It bridges the gap between algorithmic sparsity and the physical power states of commercial hardware, enabling unified frequency modulation that saves energy without sacrificing model performance.

\section{Discussion} \label{sec:Discussion}
\textbf{Distinguishing Sparsity Pattern.}
SparseDVFS currently treats sparsity largely as a scalar ratio. However, operators with the same sparsity ratio can have different patterns~\cite{lin2023efficient}, i.e., structured~\cite{wen2016learning,zhu2025sampleattention,yang2025lserve} (i.e., prune according to regular patterns to form a predictable sparse structure) or unstructured~\cite{jeong2025enabling,wilkinson2023register,gondimalla2023eureka,zhao2024chatopu} (i.e., random pruning to form arbitrary sparsity patterns). Unstructured sparsity often causes random memory access patterns that degrade bandwidth efficiency regardless of frequency. For future work, we will refine the offline modeler to penalize unstructured sparsity, potentially assigning higher EMC frequencies to handle the inefficient access patterns compared to structured sparsity.

\fakepar{Memory Access Awareness}
Our sparsity-aware method in SparseDVFS relys on the compute reduction aspect of sparsity. However, memory access overhead, specifically the latency of DRAM row activation, remains a critical factor. Future improvements could integrate real-time memory controller performance counters~\cite{kim2023moca,tsai2026page,liu2025systematic,li2023pond,woo2023scalable} into the runtime partitioner. This would allow the system to dynamically adjust the   parameter based on bus contention, rather than relying solely on static profiling.

\fakepar{Hardware Feature Adaptation}
While effective on the current hardware, the offline modeler requires device-specific profiling. A promising improvement is transfer learning~\cite{zhuang2020comprehensive,pan2009survey,ying2018transfer,neyshabur2020being}. This method would allow the V/F mapping learned on one device (e.g., NVIDIA Jetson Orin Nano) to be transferred to another (e.g., Google Edge TPU) by adapting to the new device's specific DVFS voltage steps and thermal characteristics without a full re-profiling run. This addresses the limitation of ignoring heterogeneous hardware features.

\section{Conclusion}\label{conclusion}
This paper introduces SparseDVFS, a block-level DVFS for energy efficiency edge inference. By elevating operator sparsity to a first-class metric, SparseDVFS bridges the gap between operators and the power states of commercial hardware. The block-level aggregation strategy in SparseDVFS successfully captures intra-inference computation, while mitigating the non-negligible DVFS switching latency.


\bibliographystyle{ACM-Reference-Format}
\bibliography{ref}

\end{document}